\documentclass{article} % For LaTeX2e
\usepackage{iclr2026_conference,times}

\usepackage{amsmath,amsfonts,bm}

\def\eqref#1{equation~\ref{#1}}
\def\1{\bm{1}}

\DeclareMathAlphabet{\mathsfit}{\encodingdefault}{\sfdefault}{m}{sl}
\SetMathAlphabet{\mathsfit}{bold}{\encodingdefault}{\sfdefault}{bx}{n}

\usepackage[utf8]{inputenc} % UTF-8 input encoding
\usepackage{hyperref}
\usepackage{url}
\usepackage{graphicx} % Required for inserting images
\usepackage{float}
\usepackage{enumitem}
\usepackage{booktabs} % Required for professional table formatting
\usepackage{fvextra} % Extended fancyvrb with UTF-8 support and line breaking
\DefineVerbatimEnvironment{Verbatim}{Verbatim}{breaklines=true,breakanywhere=true}

\title{How Modality Shapes Perception and Reasoning: A Study of Error Propagation in ARC-AGI}

\author{Bo Wen  \\
Hogarthian Technologies, LLC \\
\texttt{wenboown@gmail.com} \\
\And
Chen Wang \\
IBM Research \\
\texttt{Chen.Wang1@ibm.com} \\
\AND
Erhan Bilal \\
Hogarthian Technologies, LLC \\
\texttt{bilal.erhan@gmail.com}
}

\iclrfinalcopy % Uncomment for camera-ready version, but NOT for submission.

\setcitestyle{square,numbers,comma}

\renewcommand{\headrulewidth}{0pt} % Remove header rule
\renewcommand{\footrulewidth}{0pt} % Remove footer rule

\makeatletter
\renewcommand{\AddToShipoutPicture}[1]{}
\makeatother

\begin{document}

\maketitle

\fancyhead{} % Clear all headers

\begin{abstract}
    ARC-AGI and ARC-AGI-2 measure generalization-through-composition on small color-quantized grids, and their prize competitions make progress on these harder held-out tasks a meaningful proxy for systematic generalization. 
    Recent instruction-first systems translate grids into concise natural-language or DSL rules executed in generate--execute--select loops, yet we lack a principled account of how encodings shape model perception and how to separate instruction errors from execution errors. 
    We hypothesize that modality imposes perceptual bottlenecks---text flattens 2D structure into 1D tokens while images preserve layout but can introduce patch-size aliasing---thereby shaping which grid features are reliably perceived. 
    To test this, we isolate perception from reasoning across nine text and image modalities using a weighted set-disagreement metric and a two-stage reasoning pipeline, finding that structured text yields precise coordinates on sparse features, images capture 2D shapes yet are resolution-sensitive, and combining them improves execution (about 8 perception points; about 0.20 median similarity). 
    Overall, aligning representations with transformer inductive biases and enabling cross-validation between text and image yields more accurate instructions and more reliable execution without changing the underlying model.
\end{abstract}

% ============================================================================
% BRIEF DESCRIPTION FOR SUBMISSION
% ============================================================================
% This brief description can be used for the Kaggle paper submission form.
%
% We present a systematic study of how different input modalities (text vs. 
% image encodings) affect transformer-based models' perception and reasoning 
% on ARC-AGI tasks. Our approach isolates perception from reasoning using a 
% two-stage pipeline: (1) a perception task that measures how accurately models 
% identify and locate visual features across nine different encoding modalities, 
% and (2) a reasoning task that evaluates how modality choice affects instruction 
% generation and execution accuracy. We find that text encodings (JSON, ASCII) 
% excel at precise coordinate identification for sparse features, while image 
% encodings capture 2D spatial relationships but suffer from patch-size-dependent 
% aliasing effects. Critically, combining complementary modalities enables 
% cross-validation that improves both perception accuracy (by ~8 points) and 
% execution similarity (by ~0.20 median improvement) without changing the 
% underlying model. Our work provides actionable guidance for selecting context 
% encodings in ARC-solving systems and demonstrates that aligning representations 
% with transformer inductive biases significantly improves systematic 
% generalization.
% ============================================================================

\section{Introduction}
ARC-AGI and ARC-AGI-2 are benchmarks for measuring generalization through concept composition, 
emphasizing skill-acquisition efficiency under limited priors and data \cite{chollet2019measure,chollet2025arcagi2}. 
In these tasks, small color-quantized grids must be transformed by inferring concise rules from few examples. 
The ARC Prize competitions mobilize the community to pursue interpretable, reusable abstractions rather than scale-driven brute force \cite{arcprize2024,arcprize2025announcement}. 
ARC-AGI-2 further raises difficulty and tightens held-out evaluation, explicitly stress-testing approaches that rely on exhaustive search or prompt engineering shortcuts. 
Progress on this benchmark is therefore a meaningful proxy for advancing systematic generalization in AI.

Recent top systems adopt an instruction-first paradigm that translates grids into human-readable intermediates, 
proposes concise rules, and executes them in a generate–execute–select loop \cite{arclangpublic,berman2024record536,berman2025highestarcagi,pang2025arcagi,pang2025substack}. 
Berman's systems use lightweight DSLs or short English instructions with evolutionary test-time compute, achieving state-of-the-art public results. 
Pang's approach couples multi-view grid encodings with robust program synthesis and execution, rivaling or surpassing strong proprietary baselines. 
Yet these successes leave open two gaps: first, most encodings (ASCII, JSON, images) are designed for humans, with limited understanding of how transformers actually process them; 
second, failures are hard to attribute—was the instruction wrong, or was a correct instruction mis-executed by an LLM? These gaps motivate a principled study of modality and a diagnostic separation of instruction and execution.

We investigate how transformers ``see'' ARC puzzles by contrasting text and image modalities that encode the exact same grid information. 
Text serializations flatten inherently 2D structure into 1D token sequences, burdening attention with reconstructing vertical and spatial relations. 
Vision encoders preserve 2D layout via patches and positional encodings, but introduce patch-size-dependent aliasing that can distort single-cell features. 
These modality-driven bottlenecks suggest concrete, testable hypotheses: the choice of representation will systematically shape which features are perceived reliably and which are missed. 
Understanding these perceptual effects is a prerequisite for reliable instruction generation and downstream reasoning.

Our methodology isolates perception from reasoning and introduces diagnostics to disentangle instruction quality from execution reliability. 
We define a perception task across nine modalities and score free-form descriptions against ground truth using a weighted set-disagreement metric, 
and a two-stage reasoning pipeline that evaluates instruction generation and instruction following. 
This approach builds on our previous work \cite{wen2025enhancingreasoningadaptlarge}, which demonstrated that using differences between multiple reasoning attempts as signals to drive consensus can significantly improve LLM performance on domain-specific tasks. 
Here, we extend this principle to cross-modal validation, where differences between modality-specific perceptions serve as error signals that guide the model toward more accurate internal representations. 
Empirically, structured text encodings provide precise coordinates on sparse, well-delimited features, 
while image-based views capture 2D shapes but are sensitive to resolution. 
Using both together lets the model cross-check coordinates and shapes, improving execution accuracy. 
On perception, the best representation exceeds the runner-up by roughly 8 points on our accuracy metric; 
on execution, combining modalities raises median similarity by about 0.20 compared to text-only inputs.

Taken together, our results show that representation choice is pivotal for both perception and reasoning in ARC-like tasks. 
By aligning encodings with transformer inductive biases and leveraging complementary modalities for cross-validation, agents produce more accurate instructions and more reliable executions. 
We provide actionable guidance for selecting context encodings and demonstrate that multi-modal inputs improve robustness without altering the underlying model.

\section{Background}

\subsection{What are ARC-AGI and ARC-AGI-2? Why this competition?}
The Abstraction and Reasoning Corpus (ARC) was proposed as a benchmark to measure \emph{generalization through concept composition} rather than rote pattern matching. 
Chollet frames intelligence as \emph{skill-acquisition efficiency} under limited priors and experience, emphasizing scope, generalization difficulty, and sample efficiency \cite{chollet2019measure}. 
ARC instantiates these ideas with small, diverse tasks designed to be solvable by humans using a compact set of innate priors (objects, relations, geometry), discouraging overfitting to task-specific statistics.

The ARC Prize competitions operationalize this philosophy by challenging open-source community to solve held-out ARC tasks, encouraging approaches that learn to compose simple concepts into novel solutions rather than rely on brute-force search or prompt scale \cite{arcprize2024}. 
ARC-AGI-2 extends the original benchmark with harder tasks and stronger held-out evaluations, explicitly stress-testing methods that depend on exhaustive search, data leakage, or heavy prompt engineering. The stated goal is progress toward agents that can rapidly acquire new skills and generalize across tasks via interpretable, reusable abstractions.

\subsection{What does an ARC puzzle look like?}
\label{subsec:challenge-selection}
Each task consists of one or more \textbf{training} input–output grid pairs and one or more \textbf{test} inputs. 
Grids are small color-quantized images (integers 0–9) where the learner must infer the underlying rule that maps inputs to outputs and then apply that rule to produce the missing test outputs. 
Rules commonly involve object detection, symmetry, copying, filtering, counting, or simple geometry, often composed in novel ways.

Below are five tasks from Public Evaluation v2 Set (ARC2 hard) that cover a range of difficulties. 
The first training example (input–output pair) for each challenge is shown alongside its description below, which we use in our perception experiments (see Section~\ref{subsec:modality-results}). Each challenge presents distinct spatial reasoning requirements that test different aspects of a transformer's perception and abstract-reasoning:

\begin{itemize}[leftmargin=0em]
    \item \begin{minipage}[t]{0.6\textwidth}
        \textbf{\texttt{136b0064} (easy)}: This challenge features a yellow vertical divider (column H) separating a structured left panel from a mostly empty right panel. The left panel contains eight distinct colored shapes arranged in four rows within a 15×15 grid, and the puzzle requires constructing the right side based on rules inferred from the left. Its clear spatial separation, distinct shapes, and small grid size make it relatively easy for LLMs and VLMs, as the well-defined boundaries reduce coordinate confusion.
    \end{minipage}\hfill
    \begin{minipage}[t]{0.35\textwidth}
        \vspace{0pt}
        \raggedleft
        \includegraphics[width=\textwidth]{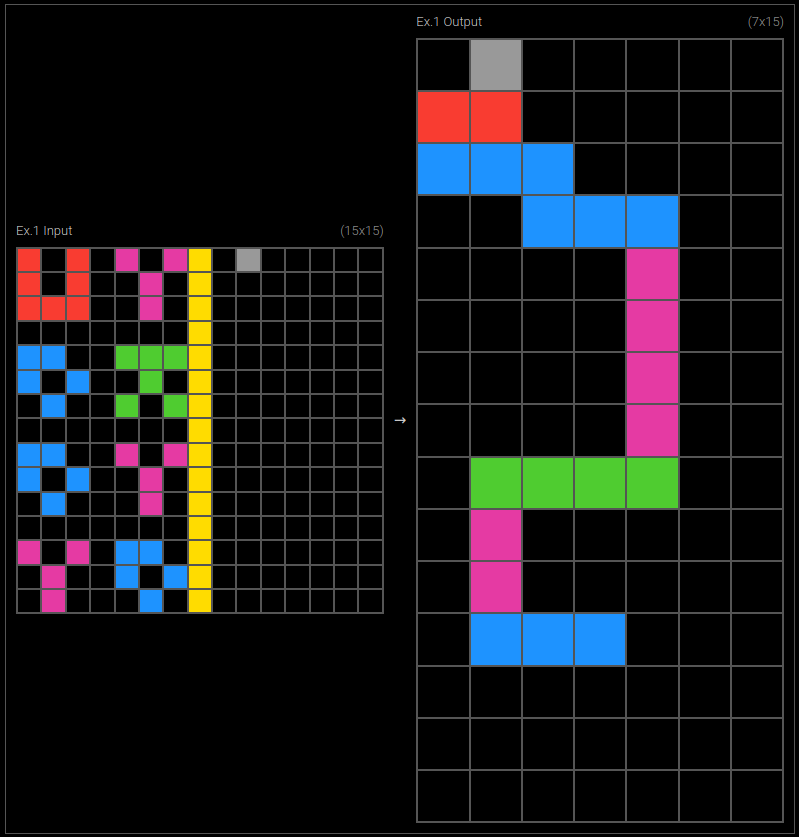}
    \end{minipage}
    
    \vspace{0.5em}
    
    \item \begin{minipage}[t]{0.6\textwidth}
        \textbf{\texttt{135a2760} (medium)}: This puzzle involves repairing repeated patterns within rectangular frames. A key difficulty is that the training examples are horizontal, while the test examples are vertical, testing the generalizability of the generated instructions. From a perception standpoint, the challenge is to pinpoint the exact locations of the repeating internal structures to determine where the translational symmetry is broken.
    \end{minipage}\hfill
    \begin{minipage}[t]{0.35\textwidth}
        \vspace{0pt}
        \raggedleft
        \includegraphics[width=\textwidth]{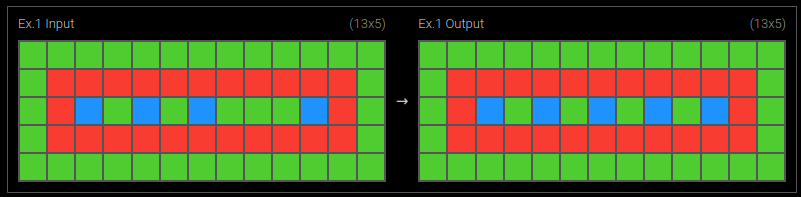}
    \end{minipage}
    
    \vspace{0.5em}
    
    \item \begin{minipage}[t]{0.6\textwidth}
        \textbf{\texttt{13e47133} (hard)}: This puzzle requires creating spiral patterns from seed dots on a canvas with a blue background and divider lines. In our experiments, a common perception failure was misreading the coordinates of the seed dots and misinterpreting the complex, multi-turn divider lines. LLMs often described these irregular areas as simple rectangles (e.g., ``K11:Q15'' or ``J11:P15''), which led to ambiguous instructions for subsequent steps.
    \end{minipage}\hfill
    \begin{minipage}[t]{0.35\textwidth}
        \vspace{0pt}
        \raggedleft
        \includegraphics[width=\textwidth]{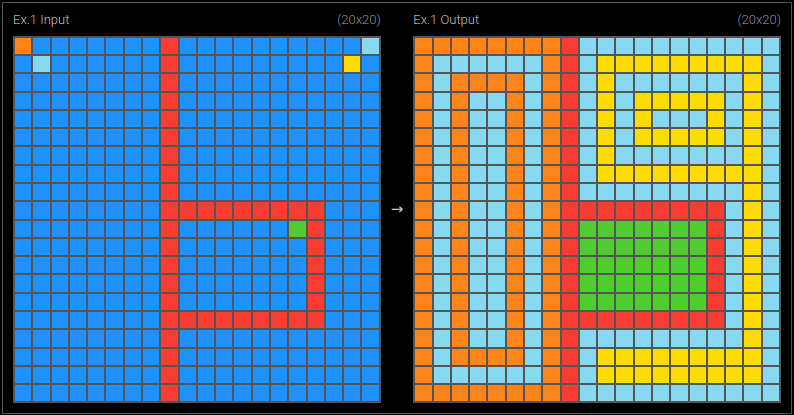}
    \end{minipage}
    
    \vspace{0.5em}
    
    \item \begin{minipage}[t]{0.6\textwidth}
        \textbf{\texttt{142ca369} (hard)}: This puzzle is perceptually simple but conceptually difficult. It features L-shaped ``laser guns'' of different colors that emit light beams, which reflect off colored ``mirrors'' and change color upon reflection. The main difficulty lies in the abstract leap required to connect the visual patterns to the concept of light reflection. The test cases add further complexity by introducing crossing light paths—a scenario not present in the training data but solvable with the real-world knowledge that light beams do not interfere when crossing.
    \end{minipage}\hfill
    \begin{minipage}[t]{0.35\textwidth}
        \vspace{0pt}
        \raggedleft
        \includegraphics[width=\textwidth]{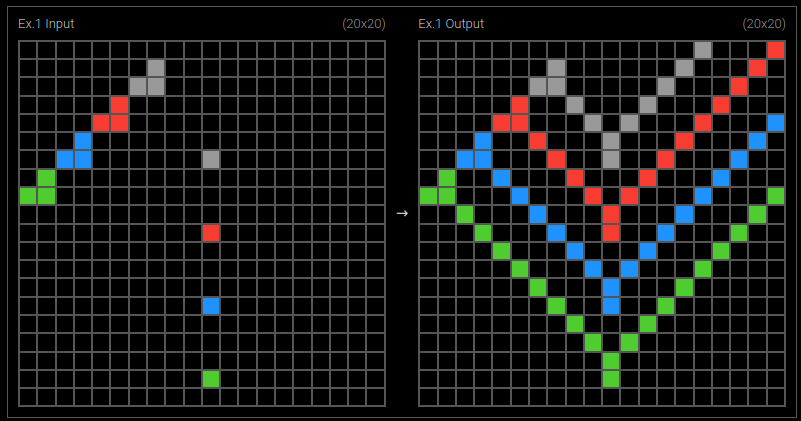}
    \end{minipage}
    
    \vspace{0.5em}
    
    \item \begin{minipage}[t]{0.6\textwidth}
        \textbf{\texttt{0934a4d8} (very hard)}: The largest and most complex challenge, this puzzle involves repairing a hole in a mosaic pattern with intricate symmetry rules. There are several difficulties: first, the pattern is off-center, requiring an adjustment in symmetry calculations. Second, while the training examples can be solved with simple x- or y-axis reflection, the test case is designed such that a standard reflection places the matching area partially outside the visible grid. To solve this, the agent must presume the puzzle is solvable—an external clue\footnote{All puzzles in the ARC Prize benchmarks are rigorously validated to be solvable by at least two humans within two attempts, providing a strong prior that a solution exists \cite{arcprize2025announcement}.}—and re-examine the training data for a hidden rule. This deeper analysis reveals that different regions of the canvas follow different symmetries: while the center and corners use standard reflection, a diamond-shaped area between them follows an additional 90-degree rotational symmetry. This rule is consistent across all examples, making it a ``universal'' constant for the challenge. The hole in the test case falls within this rotationally symmetric area, and the corresponding matching region is within the viewport, providing a high-confidence solution.
    \end{minipage}\hfill
    \begin{minipage}[t]{0.35\textwidth}
        \vspace{0pt}
        \raggedleft
        \includegraphics[width=\textwidth]{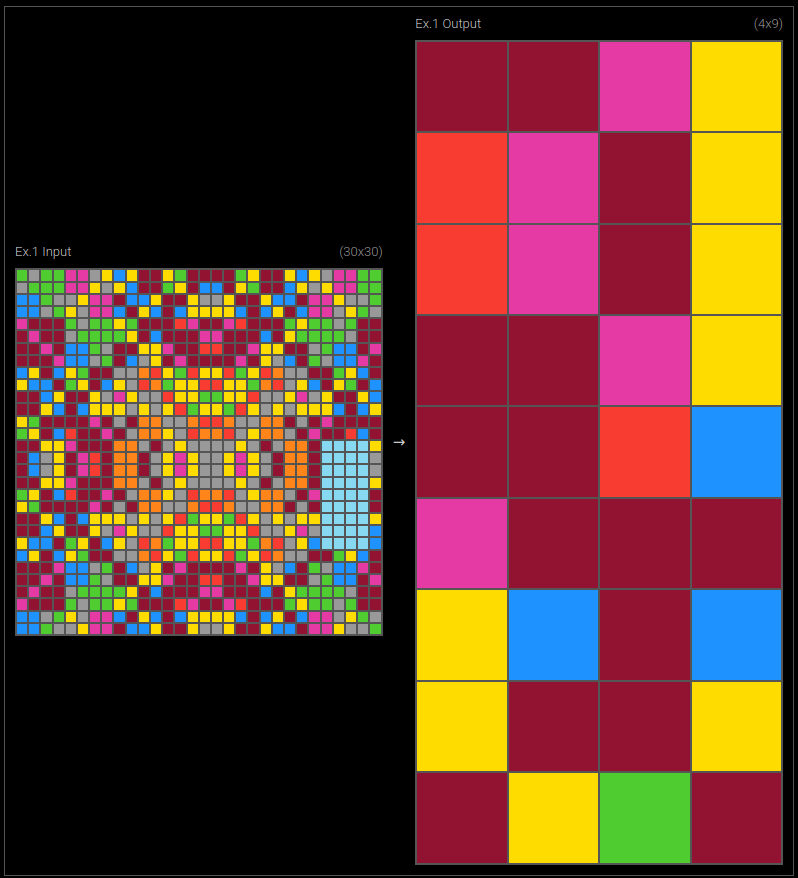}
    \end{minipage}
\end{itemize}

Readers can view these challenges interactively at the ARC Prize website by replacing the challenge ID in the URL (e.g., \url{https://arcprize.org/play?task=142ca369}).

\subsection{Instruction-centric agents on the leaderboard}
Recent top systems follow an \emph{instruction-first} paradigm:
\begin{itemize}[leftmargin=2em]
    \item Encode grids (ASCII/JSON and sometimes images), 
    propose concise natural-language \emph{instructions} (or DSL-like programs), 
    execute them with a deterministic engine, 
    and select candidates that fit all training pairs.
    \item 2024: Berman introduces an evolutionary test-time compute scheme. 
    Agents self-refine and vote over natural-language instructions executed on grids. 
    The system is ASCII-first with optional images. 
    It achieves strong held-out performance via generate–execute–select loops \cite{berman2024record536}.
    \item 2025: ``Swap Python for English''—replace Python programs with short English instructions. 
    The LLM interprets and \emph{executes} these instructions directly, removing the separate Python engine. 
    This tighter coupling plus aggressive test-time exploration improves results \cite{berman2025highestarcagi}. 
    Related public tooling includes ARC Lang \cite{arclangpublic}, a DSL for ARC puzzles. 
\end{itemize}

In this paradigm, an \textit{instruction} is a concise natural-language (or DSL) rule that specifies how to transform an input grid into an output grid, 
while \textit{execution} is the process of applying that rule to produce the output. 
It is important to note that execution is not always fully deterministic: 
when an LLM directly interprets instructions to generate output grids, it may hallucinate even when given perfectly accurate instructions. 
Similarly, when LLMs generate Python programs (as in the No.~2 approach on the leaderboard \cite{pang2025arcagi,pang2025substack}), 
although Python execution itself is deterministic, the code generation step remains prone to LLM hallucination and errors.
This distinction between instruction generation and execution is valuable for several reasons: 
it aligns with the generate–execute–select loop \cite{berman2024record536} used by top leaderboard systems, 
enables precise validation against training pairs, 
and cleanly separates failure modes (incorrect instruction vs.\ correct instruction misapplied), 
thereby supporting more scalable search strategies and effective self-correction mechanisms.

\section{How Transformers ``See'' ARC Puzzles: A Study in Modality}
The first step in solving an ARC puzzle is representing it in a format a model can understand. This choice is critical, as it can create perceptual bottlenecks that hide information needed to solve the task. 
This observation is consistent with prior ARC literature emphasizing the centrality of representation to reasoning performance \cite{chollet2019measure}.
Standard Large Language Models (LLMs) are fundamentally text-based, processing information as a 1D sequence of tokens with 1D positional encodings \cite{vaswani2017attention}. This sequential nature is a potential mismatch for ARC puzzles, which are inherently 2D spatial reasoning tasks. Reconstructing 2D relationships from a flattened 1D text sequence places a heavy burden on the transformer's attention mechanism. 
In leading instruction-first ARC systems, grids are typically serialized to plain-text for LLM consumption, sometimes augmented with images for vision-capable models. For example, Berman's ARC Lang exposes grids as human-readable ASCII matrices (space-separated digits) and optionally as base64-encoded images to aid perception during instruction generation and execution \cite{arclangpublic,berman2024record536}. Pang similarly provides multiple synchronized views of each grid (ASCII with a pipe separator, structured Python lists, dimensions, and base64 images), then uses the LLM-generated programs to transform grids \cite{pang2025arcagi}. These encodings are designed for clarity and token efficiency, but they inherit the 1D, sequential processing bias of text models that we study empirically below.
Standard transformers use \textbf{full attention}, where every token attends to every other token, leading to quadratic complexity with sequence length. This makes it computationally expensive to identify vertical patterns between tokens that are far apart in the flattened sequence. 
While \textbf{linear attention} variants reduce complexity \cite{katharopoulos2020linear,choromanski2021performer,beltagy2020longformer}, they approximate pairwise interactions and can degrade fine-grained relational precision—problematic for ARC tasks that hinge on exact local geometry and single-cell differences. Rather than rely on approximate attention for 1D text, we therefore explore vision encoders that operate natively in 2D.

In contrast, Vision-Language Models (VLMs) are designed to handle 2D data. They use patch-based tokenization, dividing an image into a grid of patches \cite{dosovitskiy2020image}. 
VLMs and vision transformers also incorporate 2D positional encodings and hierarchical windows to preserve spatial structure \cite{carion2020detr,liu2021swin}.
Crucially, they augment these patch embeddings with 2D positional encodings (x and y coordinates), allowing the self-attention mechanism to operate on a native 2D representation. This enables a true \textbf{global attention} where every patch can directly assess its relationship with every other patch across the entire grid, preserving spatial relationships without the reconstruction burden faced by LLMs. 
However, this process introduces its own challenges, such as sensitivity to the ratio of object size to patch size and potential information loss during patching, as the model's reasoning is more effective between patches than within them.
As Li notes, current multimodal LLMs typically tokenize data into 1D or 2D sequences, which makes simple spatial tasks unnecessarily difficult \cite{li2025spatialintelligence}, highlighting the fundamental challenge of representing spatial information in transformer architectures.

This fundamental difference in processing between 1D text-based models and 2D vision models inspired our central research question: how does the choice of data modality affect a transformer's ability to ``see'' and reason about ARC puzzles?
We formulate two hypotheses regarding how modality choice affects model performance:

\begin{itemize}[leftmargin=2em]
    \item \textbf{Hypothesis 1 (Perception):} The choice of modality will affect the model's ability to perceive the features in the grid. (This can break down into 4 sub-hypotheses, see Appendix \ref{appendix:modality-hypotheses})
    
    \item \textbf{Hypothesis 2 (Reasoning):} The choice of modality will affect the model's ability to generate a correct instruction that transforms the input grid into the output grid. Models that perceive spatial features more accurately will have a higher chance to generate more precise transformation rules.
\end{itemize}

See appendix \ref{appendix:modality-hypotheses} for more details.

To investigate this, we designed a modality-vision experiment to empirically test what a model perceives from different puzzle encodings.

\subsection{Modality Definitions}
\label{subsec:modality-definitions}
To understand how transformers process ARC puzzles, we systematically tested nine distinct input modalities across text and image formats. Each modality encodes the same grid information but presents it differently, affecting how the model tokenizes and attends to spatial patterns.

\paragraph{Text Modalities.}
\begin{itemize}[leftmargin=2em]
    \item \textbf{row\_only}: Row-wise text format where each row is presented as a prefix label followed by concatenated color values (no spaces, brackets, or commas). Rows are separated by double newlines. Example for a 3×10 grid:
    \begin{Verbatim}
R1: 1234567890\nR2: 1004567890\nR3: 1894567890\n...
    \end{Verbatim}
    This format emphasizes horizontal patterns but flattens the 2D structure into sequential rows.
    
    \item \textbf{col\_only}: Column-wise text format using spreadsheet column labels (A--Z, AA--AD) followed by concatenated color values. Columns are separated by double newlines. Example for a 3×10 grid:
    \begin{Verbatim}
A: 111\nB: 208\nC: 309\nD: 444\nE: 555\nF: 666\nG: 777\n...
    \end{Verbatim}
    This format emphasizes vertical patterns and naturally transposes row/column semantics (column A corresponds to the first column of the grid).
    
    \item \textbf{ascii}: Space-separated integers per row\footnote{In our experiment, we used spaces as separators. Pang's implementation uses ``\textbar'' (pipe) as the separator \cite{pang2025arcagi}, but according to our tokenization theory, this should make no difference. However, researchers seeking more rigorous conclusions are welcome to verify this using our repository \cite{arcagi2024repository}.}, similar to common ARC-AGI text encodings. Rows are separated by single newlines. Example for a 3×10 grid:
    \begin{Verbatim}
1 2 3 4 5 6 7 8 9 0
1 0 0 4 5 6 7 8 9 0
1 8 9 4 5 6 7 8 9 0
    \end{Verbatim}
    This format preserves explicit cell spacing, making individual cell values easier to parse but requiring more tokens.
    
    \item \textbf{json}: Raw JSON array of arrays (list of lists), matching the native ARC-AGI data structure. Example for a 3×10 grid:
    \begin{Verbatim}
[[1,2,3,4,5,6,7,8,9,0],[1,0,0,4,5,6,7,8,9,0],[1,8,9,4,5,6,7,8,9,0]]
    \end{Verbatim}
    This format is compact and preserves exact structure but requires parsing nested brackets and commas.
\end{itemize}

\begin{figure}[h]
    \centering
    \includegraphics[width=0.6\textwidth]{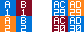}
    \caption{Example of image modality representation showing 2×2 grid regions from the upper left (columns A--B, rows 1--2) and lower right (columns AC--AD, rows 29--30) corners of a 30×30 grid from challenge \texttt{0934a4d8}. Each cell displays its spreadsheet coordinate label (e.g., ``A1'', ``B2'', ``AC29'') along with its color value. The coordinate labels use a vertical layout to fit within the cell boundaries, enabling precise spatial grounding for vision models. This example uses the \texttt{image\_16x16} modality (16×16 pixels per cell).}
    \label{fig:grid_examples}
\end{figure}

\paragraph{Image Modalities.}
All image modalities render the grid as a PNG image where each cell is a colored square. Non-empty cells (color value $\neq$ 0) include pixel-perfect spreadsheet coordinate labels (e.g., ``A1'', ``B2'', ``P11'') rendered inside the cell. The coordinate labels use a vertical layout to ensure they fit. See Figure~\ref{fig:grid_examples} for an example.

\begin{itemize}[leftmargin=2em]
    \item \textbf{image\_14x14}: 14×14 pixels per cell (minimum resolution for labels).
    
    \item \textbf{image\_15x15}: 15×15 pixels per cell.
    
    \item \textbf{image\_16x16}: 16×16 pixels per cell (default).
    
    \item \textbf{image\_17x17}: 17×17 pixels per cell.
    
    \item \textbf{image\_768x768}: Variable resolution optimized for vision models with 768×768 patch tiling (e.g., Gemini). For a 20×20 grid, this yields approximately 38×38 pixels per cell, providing high-fidelity detail. This modality is designed to align with vision encoder patch boundaries.
\end{itemize}

\section{Methodology}
\subsection{LLM's Perception of Important Visual Features}
\label{subsec:perception-methodology}

We designed a task that isolates perception from reasoning, to test our hypothesis that modality choice affects feature perception. This allows us to measure how accurately a model identifies and locates visual features when presented with the same grid in different formats. In the future, we plan to use mechanistic interpretability tools to inspect the internal representations of multimodal inputs.

For our experiments, we focused on five ARC challenges covering a range of difficulties (see Section~\ref{subsec:challenge-selection}). This selection allows us to run small controlled experiments while still covering different aspects of ARC challenges. For each task, we encoded the \textbf{input grid of the first training example} into our nine modalities and queried multi-modal LLMs (primarily Gemini 2.5 Pro, leveraging Google Cloud's free trial credits, with plans to test open-source models in future work) to produce detailed, structured descriptions of the grid. These descriptions included objects, shapes, precise locations, colors, and relationships. We then used a systematic workflow to verify these descriptions against ground truth to measure perception accuracy, focusing on single-pixel objects to isolate coordinate identification. All experimental data, verification workflows, and analysis reports are available in our repository \cite{arcagi2024repository}.

\paragraph{Perception Task.}
Our perception task prompts the model to act as a vision expert and describe the grid's contents in detail. We provide a system prompt (see Appendix~\ref{appendix:perception-prompt}) requesting information on objects, locations (using spreadsheet notation), colors, spatial relationships, patterns, and grid structure. For each ARC challenge, we encode the \textbf{input grid of first training example} into various modalities (e.g., ASCII, JSON, image resolutions) and use each as input for the model to generate a free-form text description. \footnote{Our implementation is available at \url{https://github.com/hogarthian/AIF-ARC-AGI-public/tree/main/perception}}

\paragraph{Example Perception Outputs.}
The model's descriptions vary significantly in detail and accuracy across modalities. For instance, when describing challenge \texttt{13e47133} in ASCII format, the model correctly identifies all five isolated single pixels (orange at A1, teal at B2 and T1, yellow at S2, green at P11) and describes the red structure as a ``Number 4'' shape with precise coordinate ranges. In contrast, the row-only format description correctly identifies four of the five pixels but incorrectly locates the green pixel at Q11 instead of P11 (off by one column). Image modalities with coordinate labels generally provide the most detailed spatial descriptions, explicitly referencing the coordinate overlay to ground their observations.

\paragraph{Ground Truth Features.}
For each challenge, we define ground truth features that are critical for solving the puzzle. These features serve as the basis for evaluating perception accuracy across modalities. (Remember that we are only evaluating the \textbf{input grid of first training example}.) For \texttt{13e47133}, the ground truth includes five isolated colored dots, a complex multi-turn divider line, and a blue background. For \texttt{135a2760}, we focus on nested green rectangular frames, nested red rectangular frames, and blue dots positioned at specific locations. For \texttt{142ca369}, the ground truth consists of four L-shaped objects of different colors and several isolated single dots. For \texttt{136b0064}, we evaluate identification of eight distinct colored shapes arranged in four rows, a yellow vertical divider spanning the full column H, and a grey dot. For \texttt{0934a4d8}, the critical feature is a large connected block (the ``hole'' in the mosaic pattern) that must be correctly identified; while the challenge contains many scattered colored dots throughout the mosaic, attempting to enumerate all dots is counterproductive—the correct approach is to identify symmetry patterns rather than listing individual pixels.

\paragraph{LLM-Powered Analysis for Accuracy Scoring.}
To quantify perception accuracy from the free-form text descriptions, we developed an LLM-powered analysis pipeline. 
We extract ground truth coordinates directly from the raw grid data, then use an LLM to parse each generated description and extract claimed features with their coordinates and colors. 
This approach handles the high variance in natural language descriptions better than regex matching. 
For each modality and challenge, we compare the LLM-extracted claims against ground truth to identify missing features, hallucinations, and coordinate accuracy.
To capture both omissions and hallucinations in a size-invariant way while emphasizing task-critical primitives, 
we define accuracy via normalized point-level set disagreement and aggregate it with per-feature weights.
We calculate accuracy scores using a weighted per-feature similarity approach: 
for each ground truth feature (dots, frames, blocks, or complex objects), 
we compute an individual similarity score starting at 100 and subtracting penalties proportional to discrepancies between the model's claims and ground truth. 

For a ground truth feature $F_{\mathrm{gt}}$ (represented as a set of coordinate points) and a claimed feature $F_{\mathrm{claimed}}$ (similarly represented), we define the discrepancy measure as:
\begin{equation}
|\mathrm{diff}| = \frac{|(F_{\mathrm{gt}} \setminus F_{\mathrm{claimed}}) \cup (F_{\mathrm{claimed}} \setminus F_{\mathrm{gt}})|}{|F_{\mathrm{gt}}|}.
\end{equation}
where the numerator counts all non-overlapping points between the two features (points in $F_{\mathrm{gt}}$ but not in $F_{\mathrm{claimed}}$, 
plus points in $F_{\mathrm{claimed}}$ but not in $F_{\mathrm{gt}}$). 
This unified formula covers all cases: $|\mathrm{diff}| = 0$ when features match perfectly, $|\mathrm{diff}| = 1$ when the feature is completely missing ($F_{\mathrm{claimed}} = \emptyset$), and $|\mathrm{diff}| > 1$ can happen when the claimed feature includes substantial hallucinations (points outside $F_{\mathrm{gt}}$) or is significantly larger than the ground truth, $|\mathrm{diff}| < 1$ accounts for small mistakes.

For each feature, the penalty is $w_{\mathrm{feature}} \times |\mathrm{diff}|$, where $w_{\mathrm{feature}}$ is a weight assigned to that feature based on its importance for solving the puzzle. 
The final accuracy score is $100 - \sum_{\mathrm{features}} \mathrm{penalty}$. 
Feature weights are determined by human experts who assess each feature's criticality for puzzle solving—for example, in challenge \texttt{13e47133}, isolated colored dots are weighted at 15 points each, the complex divider line at 20 points, and the background color at 5 points, reflecting their relative importance. 
This weighted approach ensures that features critical for puzzle solving contribute more to the accuracy score, while still providing a balanced assessment across different feature types. 
A detailed analysis of modality-specific failure modes and scoring breakdowns, including the specific weights used for each challenge, is presented in Section~\ref{subsec:modality-results} and Appendix~\ref{appendix:perception-detail-result}.

\subsection{The Effect of Modality on LLM's Reasoning}

While perception accuracy measures what a model can ``see,'' reasoning accuracy measures how well a model can abstract patterns and clues into transformation rules, and apply them to solve puzzles. To our \textbf{Hypothesis 2 (Reasoning)}, we designed an experiment that separates instruction generation from instruction application, allowing us to examine the impact of modality on each stage of the reasoning pipeline.

\paragraph{Reasoning Task.}
Our reasoning experiment follows a two-stage pipeline inspired by \cite{berman2025highestarcagi}. First, we generate transformation instructions by prompting (See Appendix \ref{appendix:reasoning-prompt}) the model to analyze training examples and produce a structured hypothesis containing: (1) a \textit{working hypothesis} describing discovered rules and patterns, including clues needs multiple training examples to formulate(the abstraction step); (2) \textit{general instructions} specifying reusable step-by-step algorithms; (3) \textit{example instructions} showing how the general rules apply to each training example; and (4) \textit{test instructions} providing detailed steps for solving test cases (the reasoning step). The model generates these instructions using grids encoded in a specific modality (e.g., \texttt{row\_only}, \texttt{row\_col\_image}).

Second, we apply these instructions by prompting (See Appendix \ref{appendix:follow-instruction-prompt}) the model to reason through the one instructions and transform an input grid into its output grid. The reasoning prompt includes the working hypothesis and instructions, along with grids encoded in the same modality used during instruction generation. Crucially, we test two reasoning contexts: \textit{normal} (with training examples visible as reference) and \textit{reduced} (without training examples, relying solely on one ``output'' instruction). This allows us to measure whether working hypothesis included all the cross-training example clues.\footnote{Our implementation is available at \url{https://github.com/hogarthian/AIF-ARC-AGI-public/tree/main/generate-execute}}

\paragraph{Reasoning Accuracy Metric.}
Again inspired by \cite{berman2025highestarcagi}, we define reasoning accuracy as the fraction of cells that match exactly between the generated output grid and the ground-truth output grid. Formally, for grids $G_{\mathrm{expected}}$ and $G_{\mathrm{generated}}$ of dimensions $R \times C$, the similarity score is
$\mathrm{similarity} = |\{(r,c) : G_{\mathrm{expected}}[r][c] = G_{\mathrm{generated}}[r][c]\}| / (R \times C)$.
This yields a score between 0.0 (no matches) and 1.0 (perfect match). If the grids have different dimensions, we define $\mathrm{similarity}=0$. We calculate this metric for both held-out validation (leave-one-out on training examples) and test cases, allowing us to measure both generalization within the training distribution and performance on unseen test inputs.

\section{Experiments}

\subsection{Impact of Modality on LLM's Perception}
\label{subsec:modality-results}

\begin{table}[htpb]
\centering
\small
\begin{tabular}{lccccc}
\toprule
\textbf{Modality} & \textbf{13e47133} & \textbf{135a2760} & \textbf{136b0064} & \textbf{142ca369} & \textbf{0934a4d8} \\
\midrule
Row-only & 60.61 & 100.00 & 100.00 & 87.50 & 5.00 \\
Column-only & 70.00 & 100.00 & 100.00 & 100.00 & 21.67 \\
ASCII & 87.32 & 100.00 & 100.00 & 91.67 & -11.67 \\
JSON & 100.00 & 100.00 & 100.00 & 100.00 & 30.00 \\
Image 14×14 & 49.15 & 100.00 & 90.00 & 25.00 & -21.67 \\
Image 15×15 & 52.56 & 76.25 & 100.00 & 0.00 & 50.00 \\
Image 16×16 & 45.73 & 96.25 & 100.00 & 37.50 & -66.11 \\
Image 17×17 & 68.05 & 96.25 & 100.00 & 100.00 & -132.78 \\
24×24-1148 & 100.00 & 100.00 & 90.00 & 0.00 & -32.78 \\
24×24-1205 & 85.00 & 96.25 & 50.00 & 0.00 & 45.00 \\
Image 768×768 & 70.00 & 100.00 & 100.00 & 25.00 & 6.11 \\
\bottomrule
\end{tabular}
\caption{Perception accuracy scores by challenge and modality. Scores measure how accurately each modality identifies and locates ground truth features when describing the input grid of the first training example. Using a weighted per-feature similarity approach (see Section~\ref{subsec:perception-methodology}), each feature contributes to the final score according to its importance for puzzle solving, calculated as $100 - \sum_{\mathrm{features}} \mathrm{penalty}$, where penalties are $w_{\mathrm{feature}} \times |\mathrm{diff}|$ based on coordinate discrepancies (Equation~1). Feature weights $w_{\mathrm{feature}}$ are assigned by human experts based on each feature's criticality for solving that particular challenge. Negative scores indicate severe misidentification where penalties exceeded the base score. Detailed per-challenge analysis including specific feature weights is available in Appendix~\ref{appendix:perception-detail-result}.}
\label{tab:modality-feature-accuracy}
\end{table}

Table~\ref{tab:modality-feature-accuracy} presents perception accuracy scores across five ARC challenges, 
revealing systematic differences that strongly support 
\textbf{Hypothesis 1}: modality choice fundamentally affects a transformer's ability to perceive grid features.

Our experiments demonstrate that text and image modalities exhibit complementary strengths and failure modes. 
Text modalities excel on structured, well-defined features: JSON achieves perfect accuracy (100\%) on challenges \texttt{13e47133}, \texttt{135a2760}, and \texttt{142ca369}, where features are sparse but clearly delimited. 
Structured formats with explicit separators (\texttt{ascii}, \texttt{json}) outperform separator-free formats (\texttt{row\_only}) on coordinate identification tasks, achieving 80.4\% average accuracy on sparse single pixels compared to 76.4\% for \texttt{row\_only}.
However, text modalities reveal a \textit{1D processing bias}: \texttt{row\_only} correctly identifies horizontal features but struggles with vertical alignment (e.g., on challenge \texttt{13e47133}, identifying Q11 instead of P11, off by 1 column), while \texttt{col\_only} excels at vertical patterns but misses horizontal features (e.g., on challenge \texttt{13e47133}, identifying B1 instead of B2 and S1 instead of S2, both off by 1 row). 
This directional bias is most evident when comparing performance across challenges: on challenge \texttt{142ca369}, \texttt{col\_only} achieves perfect accuracy (100\%) by correctly identifying all four single dots in column K, while \texttt{row\_only} achieves 87.50\% by misidentifying K7 as L7 (off by 1 column), demonstrating that serialization direction determines which spatial relationships the model can reliably reconstruct.
In practice, it is recommended to use \texttt{json} or \texttt{ascii} for challenges with sparse, clearly delimited features that require precise coordinate identification, \texttt{row\_only} for challenges dominated by horizontal patterns, and \texttt{col\_only} for challenges dominated by vertical patterns.

Image modalities provide native 2D representation with explicit coordinate labels, enabling better shape recognition for complex patterns. 
However, they introduce \textit{visual aliasing effects} from patch-based tokenization that create systematic hallucinations. 
At low resolutions (14×14–17×17 pixels per cell), multiple grid cells pack into single vision patches, causing the model to perceive single pixels as larger shapes (e.g., describing a single pixel as a 2×2 square or 1×2 rectangle). 
At high resolutions, single cells span multiple patches, leading to cross-boundary counting errors and coordinate misreadings when spreadsheet labels straddle patch boundaries. 
White coordinate annotations can also alter patch embeddings, causing color misreadings (e.g., teal cells misread as yellow). 
These aliasing effects explain why boundary accuracy varies non-linearly with resolution: only Image 15×15 and 24×24-1205 achieve perfect boundary accuracy (100\%) on dense blocks, while other resolutions show significant errors. 
The optimal resolution (24×24-1205, approximately 1.5 patches per cell for Gemini's 16×16 patch size) prevents critical features from aligning precisely with patch boundaries, reducing aliasing artifacts. \footnote{We ran the 24×24 experiment twice (24×24-1148 and 24×24-1205) and observed slight differences in results (e.g., 100\% vs. 85\% accuracy on challenge 13e47133), suggesting that future work should conduct multiple runs per configuration and test across different LLMs/VLMs to establish statistical significance for these findings.}
In short, the (often hidden) vision patch size that drives VLM attention is pivotal: 
when cell resolution misaligns with patch size, patch-based tokenization induces aliasing and boundary errors, 
whereas patch-size-aware settings (approximately 1.5 patches per cell) yield more reliable pattern recognition.

The detailed failure mode analysis in Appendix~\ref{appendix:perception-detail-result} reveals that most image modality errors stem from these visual aliasing mechanisms rather than fundamental limitations of 2D representation. 
The systematic differences across challenges further demonstrate modality-specific weaknesses: on challenge \texttt{136b0064}, which contains eight distinct colored shapes arranged in four rows within a 15×15 grid, all text modalities (Row-only, Column-only, ASCII, JSON) achieve perfect accuracy (100\%), correctly identifying all eight shapes, the yellow vertical divider (H1-H15), and the grey dot (J1) with precise coordinates. 
Most image modalities also achieve perfect or near-perfect accuracy: Image 15×15, 16×16, 17×17, and 768×768 all achieve 100\%, while Image 14×14 and 24×24-1148 achieve 90\% due to minor hallucinations (e.g., Image 14×14 hallucinated the grey dot as a 1×2 rectangle instead of a single pixel). 
However, 24×24-1205 achieves only 50\% accuracy, making multiple coordinate errors: extending the Red U-shape 2 rows beyond ground truth (A1-C5 instead of A1-C3), misidentifying colors in the Pink Y-shape at E1-G3, and missing rows in Blue Q-shapes (claiming A6-C7 instead of A5-C7, and A10-C11 instead of A9-C11). 
This demonstrates that while most modalities handle this structured challenge well, patch alignment issues can still cause significant errors even at optimal resolutions. 
Similarly, on challenge \texttt{0934a4d8}, which requires identifying a large block within a complex mosaic pattern, only Image 15×15, 24x24-1205 and JSON achieve reasonable accuracy (50.00\%, 45.00\% and 30.00\%, respectively), 
while most other modalities struggle or fail completely. 
Note that scores for this challenge include symmetry analysis bonuses: all text modalities (Row-only, Column-only, ASCII, JSON) received +10 points each for detecting symmetry patterns (bilateral, vertical, or horizontal), 
while Image 15×15 received +5 points for partial symmetry detection, and all other image modalities received 0 points for failing to detect symmetry. 
This bonus structure partially explains the better performance of text modalities on this challenge, as they successfully identified the mosaic's symmetry structure while most image modalities did not. 
Taken together, these results show that text-based encodings (\texttt{json}, \texttt{ascii}) deliver precise coordinate parsing and symmetry cues but retain 1D serialization biases; 
\texttt{row\_only}/\texttt{col\_only} sharpen this bias—strong along their respective axis yet error-prone orthogonally; 
and image encodings capture 2D shapes and spatial relations but are vulnerable to patch-size aliasing and can under-detect symmetry—confirming modality-specific perceptual bottlenecks (\textbf{Hypothesis 1}) and offering concrete guidance to match encodings to task features.\footnote{The perception accuracy difference between the best representation (JSON, average 86.00\% across all five challenges) and runner-up (Column-only, average 78.33\%) is approximately 8 points, calculated by averaging each modality's scores from Table~\ref{tab:modality-feature-accuracy} across all challenges.}

Motivated by these complementary strengths and weaknesses, we next show that combining modalities enables cross-examination that mitigates individual failure modes.
Specifically, image modalities alone do not consistently outperform text modalities in perception—and often fail to detect symmetry patterns that text identifies—but their strength emerges when paired with text to enable cross-examination of features. 
As we demonstrate in Section~\ref{subsec:multimodality-execution}, such multi-modal representations consistently outperform pure-text baselines by allowing the model to cross-validate spatial information: 
text provides precise coordinate identification and structured feature parsing, whereas image captures spatial relationships and shape patterns via global attention. 
This cross-examination yields a more accurate and complete internal grid representation, which in turn strengthens reasoning and improves execution accuracy (Figure~\ref{fig:cross_order_scatter_combined}).

\subsection{Multi-modality Improves Execution Accuracy}
\label{subsec:multimodality-execution}

Building on the perception findings from Section~\ref{subsec:modality-results}, we now examine how modality choice affects the end-to-end reasoning pipeline by evaluating how different modality combinations affect execution accuracy on challenge \texttt{13e47133}.\footnote{In this experiment, image modalities use 16×16 pixels per cell resolution. Each modality was evaluated with multiple runs (20–40 data points per modality, combining held-one-out validation across 3 training examples and 2 test cases, with some modalities tested under both ascending and descending example orders). We conducted additional experiments exploring the effect of different orders when providing training examples in the prompt (ascending vs. descending), but found that order does not significantly change results. While we do not elaborate on these order effects in this paper, these multiple runs provide statistical support for the conclusions in this section, as each modality was evaluated across numerous test cases and configurations.} 
Our core finding is that \textit{combined modalities improve perception accuracy and completeness, and thus improve reasoning results}: 
when the model can cross-examine features across multiple modalities, it constructs a more accurate and complete internal representation of the grid, 
which in turn enables more precise transformation rule generation and higher execution accuracy, supporting \textbf{Hypothesis 2 (Reasoning)}.

\begin{figure}[h!]
    \centering
    \includegraphics[width=\textwidth]{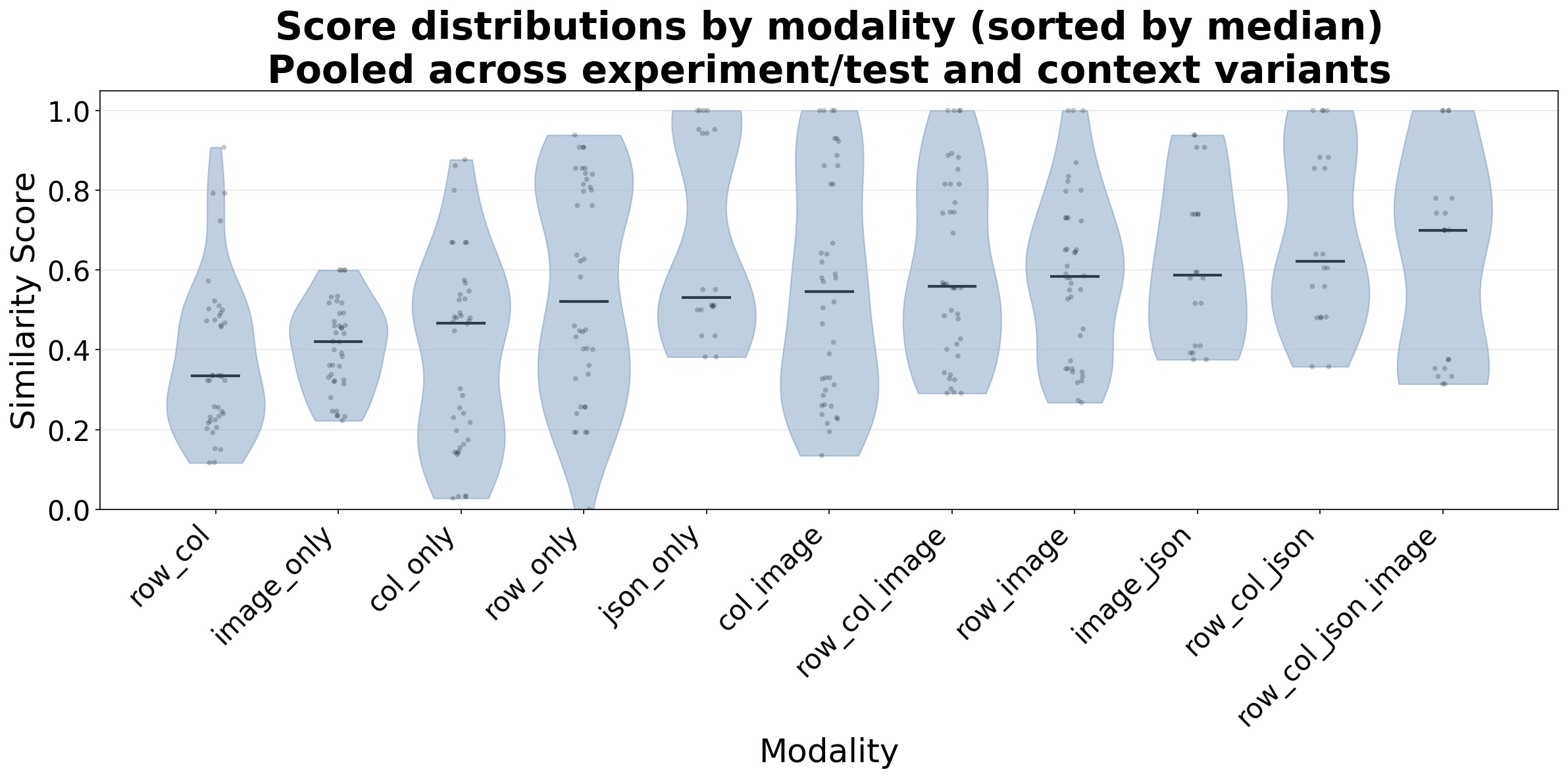}
    \caption{Score distributions by modality (sorted by median) pooled across experiment/test and context variants on challenge \texttt{13e47133}. Each violin plot shows the probability density of similarity scores, with horizontal lines indicating medians and individual gray dots representing specific data points. Modalities are sorted left-to-right by increasing median score. The plot includes scores from all three training examples (via held-one-out validation) and two test cases, revealing systematic differences in execution accuracy across modality combinations.}
    \label{fig:cross_order_scatter_combined}
\end{figure}

Figure~\ref{fig:cross_order_scatter_combined} reveals a clear pattern that directly supports our hypothesis: 
\textit{combining multiple modalities consistently improves execution accuracy by enabling cross-examination of features}. 
The progression from left to right shows single-modality encodings (\texttt{row\_only}, \texttt{col\_only}, \texttt{image\_only}, \texttt{json\_only}) achieving lower median scores, 
while multi-modal combinations (\texttt{row\_col\_json}, \texttt{row\_col\_json\_image}) achieve the highest median scores. 
The best-performing modality, \texttt{row\_col\_json\_image}, which combines all available input representations, achieves a median similarity score of approximately 0.69—nearly double that of the worst-performing single modalities. 
This improvement stems from the model's ability to cross-validate spatial information across modalities: when text modalities provide precise coordinate identification (as shown in Section~\ref{subsec:modality-results}), image modalities can confirm spatial relationships, and vice versa. 
This cross-examination process allows the model to construct a more accurate and complete internal representation, reducing the propagation of perception errors into reasoning failures.

The critical link between perception accuracy and execution performance is evident when examining \texttt{image\_only} (16×16 pixels per cell), 
which achieves a median score of approximately 0.42. 
As documented in Section~\ref{subsec:modality-results} and Appendix~\ref{appendix:perception-detail-result}, \texttt{image\_16x16} achieved only 45.73\% perception accuracy on challenge \texttt{13e47133}, 
frequently misidentifying seed dot coordinates and misinterpreting the complex divider line structure due to visual aliasing effects from patch-based tokenization. 
These perception errors directly propagate to reasoning failures: incorrect seed locations lead to wrong spiral patterns, and misread divider boundaries produce ambiguous transformation rules. 
This demonstrates that \textit{incomplete or inaccurate perception creates a weak foundation for reasoning}, leading to systematic execution failures.

In contrast, multi-modal combinations compensate for individual modality weaknesses by enabling complementary feature detection. 
As established in Section~\ref{subsec:modality-results}, text modalities excel at precise coordinate identification and structured feature parsing (e.g., JSON achieving 100\% accuracy on sparse single pixels), 
while image modalities capture spatial relationships and shape patterns through native 2D representation. 
When combined, the model can leverage text modalities to accurately identify seed coordinates while using image modalities to verify spatial relationships and shape patterns—creating a more complete and accurate internal representation that supports better reasoning. 
This explains why combinations like \texttt{row\_col\_image} and \texttt{row\_col\_json\_image} consistently outperform pure text modalities (\texttt{row\_only}, \texttt{col\_only}, \texttt{json\_only}), even though image modalities alone struggle with precise coordinate identification.

However, not all combinations provide genuine complementarity. 
The \texttt{row\_col} modality (combining row-only and column-only text formats) performs worse than either format alone, achieving the lowest median score (approximately 0.33). 
This counterintuitive result suggests that simply concatenating two 1D serializations does not enable effective cross-examination—both formats share the same fundamental limitation of sequential processing, and their combination may introduce conflicting biases without adding new spatial cues. 
However, we note that one hypothesis generated for \texttt{row\_col} exhibited strong hallucination, which affected approximately half of the follow-instruction output results and likely contributed to this low score. 
Future work should conduct multiple reruns to determine whether this poor performance reflects a systematic limitation of the \texttt{row\_col} combination or an outlier effect from a single problematic hypothesis. 
The key distinction is that combining text with image modalities (\texttt{row\_col\_image}, \texttt{row\_col\_json\_image}) provides genuine complementarity by pairing sequential processing with native 2D representation, enabling true cross-modal validation.

The vertical spread within each violin plot reflects variability across different test cases, further supporting the perception-reasoning link. 
Our analysis pools similarity scores from all three training examples (evaluated via held-one-out cross-validation) and two test cases. 
Inspection of individual results reveals systematic patterns that mirror perception difficulty: the third training example consistently achieves the highest scores across most modalities, 
due to its simple canvas division with clear, regular boundaries that are easier to perceive and reason about. 
The first training example follows, with common failures occurring on the right side's irregular channel—a complex multi-turn divider line that challenges both perception (as documented in Appendix~\ref{appendix:perception-detail-result}) and reasoning. 
The second training example typically achieves the worst scores in held-one-out validation, primarily due to yellow seed dots on a yellow background, creating color confusion that makes it difficult to distinguish seeds from the background—a perception challenge that directly undermines the model's ability to formulate correct transformation rules. 
The two test cases perform worst overall, combining the irregular boundary challenges with significantly more seed dots, increasing both perceptual complexity and reasoning difficulty. 
This consistent correlation between perception challenges and execution failures reinforces our core thesis: \textit{accurate and complete perception provides the foundation for successful reasoning}.

The systematic improvement from single to multi-modal encodings validates \textbf{Hypothesis 2}: 
models that perceive spatial features more accurately and completely through complementary modalities construct better internal representations, 
which in turn enable more precise transformation rule generation and higher execution accuracy. For example, \texttt{row\_col\_json\_image} achieves a tight distribution concentrated at higher scores (median 0.69), with most data points above 0.3, indicating consistently high performance across different test cases. This demonstrates that when modalities enable effective cross-examination—allowing the model to verify coordinate identifications, validate spatial relationships, and confirm shape patterns—the resulting internal representation supports robust reasoning even when individual modalities have weaknesses.\footnote{The execution similarity improvement of approximately 0.20 is calculated as the difference between the best multi-modal combination (\texttt{row\_col\_json\_image}, median 0.69 from Figure~\ref{fig:cross_order_scatter_combined}) and the average median of text-only baselines (\texttt{row\_only} 0.52, \texttt{col\_only} 0.47, \texttt{json\_only} 0.54; average 0.51), yielding 0.69 - 0.51 = 0.18, rounded to 0.20.}

These findings demonstrate that modality choice significantly impacts not just perception (as shown in Section~\ref{subsec:modality-results}) but also the end-to-end abstraction and reasoning capability of transformer-based agents. 
The key mechanism is that multi-modal representations enable cross-examination of features, creating a more accurate and complete internal representation of the grid that serves as a stronger foundation for reasoning. 
This provides actionable guidance for designing ARC-solving systems: combining complementary modalities that enable effective cross-validation leads to better perception accuracy and completeness, which in turn enables more robust spatial reasoning and higher execution accuracy.

\section{Discussion}

Our experiments reveal that modality choice fundamentally shapes transformer perception and reasoning on ARC tasks through two interconnected mechanisms: \textit{perceptual bottlenecks} that limit what features a model can accurately identify, and \textit{cross-modal validation} that enables more complete internal representations. These findings have immediate practical implications for designing ARC-solving systems and broader implications for understanding how transformers process spatial information.

\subsection{Perception as a Foundation for Reasoning}

The most striking finding from our experiments is the direct link between perception accuracy and execution performance. When \texttt{image\_16x16} achieved only 45.73\% perception accuracy on challenge \texttt{13e47133}—misidentifying seed dot coordinates and misinterpreting divider line structures due to visual aliasing—these errors directly propagated to reasoning failures, yielding a median execution similarity of only 0.42. Incomplete or inaccurate perception creates a weak foundation: incorrect seed locations lead to wrong spiral patterns, and misread boundaries produce ambiguous transformation rules that fail to generalize. This demonstrates that \textit{perception errors cascade through the reasoning pipeline}, making accurate feature identification a prerequisite for successful abstraction and rule generation.

This error propagation mechanism aligns with predictive coding principles from neuroscience \cite{rao1999predictive,friston2005theory}, where hierarchical systems propagate prediction errors upward through processing layers. In our framework, perception errors at the input encoding stage propagate upward to reasoning failures, creating cascading failures that cannot be corrected downstream. The cross-modal validation mechanism we observe—where complementary modalities enable error detection and correction—mirrors how predictive coding systems use multiple sensory streams to resolve ambiguities and correct prediction errors. While we do not explicitly implement a predictive coding architecture, our findings demonstrate that designing systems with hierarchical error propagation in mind—and providing mechanisms for cross-validation—can significantly improve robustness, consistent with predictive coding's emphasis on error minimization through hierarchical feedback.

Conversely, when modalities enable accurate perception, the resulting internal representations support robust reasoning. Text modalities excel at precise coordinate identification on structured tasks (JSON achieving 100\% accuracy on challenges \texttt{13e47133}, \texttt{135a2760}, and \texttt{142ca369}), but their 1D processing bias creates systematic coordinate slips when features are misaligned with the serialization direction. Image modalities preserve 2D spatial relationships through global attention, but patch-based tokenization introduces visual aliasing that corrupts single-cell detail when resolution misaligns with patch boundaries. The optimal resolution (approximately 1.5 patches per cell) prevents critical features from aligning precisely with patch boundaries, but this sweet spot is model-specific and requires careful calibration. These modality-specific bottlenecks explain why no single encoding dominates across all challenge types: the optimal format depends on how well the serialization strategy aligns with the challenge's spatial characteristics.

\subsection{Cross-Modal Validation as a Mechanism for Improvement}

Multi-modal combinations improve execution accuracy not by simply adding more information, but by enabling \textit{cross-examination} that mitigates individual failure modes. The best-performing combination, \texttt{row\_col\_json\_image}, achieves a median similarity score of 0.69—nearly double that of single-modality encodings—by allowing the model to cross-validate spatial information: text modalities provide precise coordinate identification while image modalities confirm spatial relationships and shape patterns. This cross-examination process constructs a more accurate and complete internal representation, reducing the propagation of perception errors into reasoning failures.

However, not all combinations provide genuine complementarity. The \texttt{row\_col} modality performs worse than either format alone, suggesting that simply concatenating two 1D serializations does not enable effective cross-examination when both formats share the same fundamental limitation. The key distinction is that combining text with image modalities provides genuine complementarity by pairing sequential processing with native 2D representation, enabling true cross-modal validation. This finding has practical implications: designers should combine modalities with complementary strengths rather than simply adding more channels, and should verify that combinations enable cross-validation rather than introducing conflicting biases.

\subsection{Practical Implications for ARC-Solving Systems}

Our findings provide concrete guidance for designing ARC-solving systems. For challenges with sparse, clearly delimited features requiring precise coordinate identification, structured text formats (\texttt{json}, \texttt{ascii}) deliver the highest perception accuracy. For challenges dominated by horizontal or vertical patterns, directional serializations (\texttt{row\_only}, \texttt{col\_only}) can leverage 1D processing biases as strengths rather than weaknesses. Image modalities should be paired with patch-size-aware resolution settings (approximately 1.5 patches per cell for Gemini's 16×16 patch size) to minimize aliasing artifacts, though optimal resolution varies across VLM architectures. Most importantly, combining complementary modalities enables cross-validation that improves both perception accuracy (by roughly 8 points) and execution similarity (by approximately 0.20 over text-only baselines) without altering the underlying model.

The systematic correlation between perception challenges and execution failures across different test cases further reinforces the perception-reasoning link. Training examples with simple, regular boundaries achieve consistently higher scores, while those with irregular boundaries or color confusion create perception challenges that directly undermine reasoning. This suggests that ARC-solving systems should prioritize accurate perception through modality selection and cross-validation, as perception errors create fundamental bottlenecks that cannot be overcome through reasoning alone.

\subsection{Limitations and Future Directions}

Our experiments have several limitations that future work should address. First, our perception experiments tested individual modalities in isolation rather than combined modalities, leaving open the question of how multi-modal inputs directly affect feature identification accuracy. Second, our execution experiments focused on a single challenge (\texttt{13e47133}), and while the systematic patterns suggest generalizability, broader evaluation across diverse challenge types would strengthen our conclusions. Third, the optimal image resolution is model-specific and depends on the vision encoder's patch size, requiring empirical calibration for each VLM architecture. Fourth, we observed variability in results across multiple runs (e.g., 24×24-1148 vs. 24×24-1205), suggesting that future work should conduct multiple runs per configuration and test across different LLMs/VLMs to establish statistical significance. 

Future work should also explore mechanistic interpretability to trace how each modality is encoded in to the internal representation and whether/how attention heads actually fuse multi-modal evidence together, and how all these affects the down stream reasoning token distribution changes, providing deeper insights into the cross-examination mechanism. Additionally, broadening the model sweep to include open-source models and increasing statistical power through larger-scale experiments would strengthen the generalizability of our findings. Finally, investigating how modality choice affects other aspects of ARC-solving—such as hypothesis generation, instruction refinement, and candidate selection—could reveal additional mechanisms by which representation shapes reasoning.

\subsection{Broader Implications}

More broadly, our findings demonstrate that representation is a first-order design choice rather than a formatting detail. Aligning encodings with transformer inductive biases—and letting modalities check one another—yields more accurate instructions and more reliable execution without changing the underlying model. This suggests that careful attention to input representation can yield significant performance improvements in transformer-based agents, particularly for tasks requiring spatial reasoning. The systematic differences we observe across modalities reveal how transformer architectures process spatial information differently depending on how it is encoded, providing insights into the relationship between input format and internal representation that extend beyond ARC tasks to other domains requiring spatial reasoning. This work contributes to the broader research agenda on spatial intelligence \cite{li2025spatialintelligence}, demonstrating how modality choice affects spatial perception and reasoning in transformer-based systems.

This work fits into a broader research agenda on developing inherently safer AGI systems through principled architectures \cite{wen2025frameworkinherentlysaferagi}. The error propagation and cross-modal validation mechanisms we observe align with Active Inference principles, where prediction errors drive belief updates and multiple sensory streams enable robust perception. Our findings on how perception errors cascade through reasoning pipelines provide empirical validation for hierarchical error propagation models, while the cross-modal validation mechanism demonstrates how complementary information sources can serve as natural error correction mechanisms. Future work will explore how these modality-driven insights can inform the design of multi-agent Active Inference systems for ARC-AGI, where specialized agents with different perceptual capabilities can collaborate through error-driven consensus mechanisms similar to the cross-modal validation we observe here.

\section{Related Work}

The ARC-AGI and ARC-AGI-2 benchmarks were designed to measure generalization-by-composition and resist scale-driven brute force \cite{chollet2019measure,chollet2025arcagi2,arcprize2024,arcprize2025announcement}. 
Recent state-of-the-art systems adopt instruction-first, program-synthesis pipelines that translate grids into human-readable intermediates for an LLM to propose rules, followed by a deterministic execution engine that applies those rules to transform inputs into outputs. 
Two highly influential exemplars are Berman's ARC Lang and Pang's DreamCoder-inspired system \cite{arclangpublic,berman2024record536,pang2025arcagi}.

Berman's 2024 and 2025 systems achieved top positions on the public ARC-AGI leaderboards, emphasizing multi-agent collaboration and evolutionary test-time compute, with a lightweight DSL and ASCII/image grid presentations that are easy for LLMs to parse \cite{berman2024record536,berman2025highestarcagi}. 
Pang's approach---which couples structured multi-view grid encodings with LLM-assisted synthesis and a robust executor---has reached the second-highest ranking, and notably reports outperforming advanced proprietary models (e.g., o3/o4) on public ARC-AGI evaluations \cite{pang2025arcagi,pang2025substack}. 
These instruction-centric systems demonstrate that careful engineering of representations, verification loops, and search strategies can produce strong performance without relying on end-to-end monolithic models.

Beyond these leadership results, a broader line of work explores hybrid neuro-symbolic reasoning, tool-augmented LLMs, and scalable program synthesis for compositional tasks. 
Prior advances in transformer architectures and vision-language modeling inform how representation choices shape what models can perceive and manipulate \cite{vaswani2017attention,dosovitskiy2020image,carion2020detr,liu2021swin}. 
Recent work on spatial intelligence highlights the fundamental importance of spatial reasoning for AI systems and the limitations of current approaches that tokenize spatial data into 1D or 2D sequences \cite{li2025spatialintelligence}.
Within ARC research specifically, community systems frequently serialize grids to text (e.g., space-separated rows, JSON lists) and sometimes add image renderings to aid perception, but systematic analysis of how such modalities influence what an LLM ``sees'' has been limited.

Our work is complementary. 
Rather than proposing a new end-to-end solver, we study the perceptual bottlenecks induced by common grid encodings and quantify how different modalities affect an LLM/VLM's ability to extract spatial features critical to downstream reasoning. 
This lens helps explain why top instruction-first systems benefit from multi-view encodings and suggests principled ways to design representations that reduce coordinate confusion and improve rule discovery.

\section{Conclusion}

In this paper, we studied how input modality shapes what transformers can perceive and how those percepts propagate to instruction-first reasoning on ARC-like tasks. 
By isolating perception from execution with a weighted set-disagreement metric and a two-stage generate–execute pipeline, we show that representation is a first-order design choice rather than a formatting detail.
Text encodings (especially \texttt{json}/\texttt{ascii}) deliver precise coordinates on sparse, well-delimited features, while row- and column-only serializations sharpen complementary 1D biases; 
image views preserve 2D information and relational structure via global attention but exhibit patch-size aliasing that can corrupt single-cell detail. 
Combining modalities enables cross-examination of coordinates and shapes, improving both stages: the best representation exceeds the runner-up by roughly 8 points on perception, 
and multi-modal inputs raise median execution similarity by about 0.20 over text-only baselines. 
In practice, match encodings to task structure (use \texttt{json}/\texttt{ascii} for precise coordinates; row/col for dominant directional patterns) and pair them with patch-aware images for global spatial cues.
More broadly, aligning encodings with transformer inductive biases—and letting modalities check one another—yields shorter, more accurate instructions and more reliable execution without changing the underlying model. 
We release artifacts and protocols to facilitate replication and extensions; future work will broaden the model sweep, increase statistical power, and use mechanistic interpretability to trace how multi-modal evidence is fused during perception and reasoning.

\bibliography{references}
\bibliographystyle{unsrt}

\appendix
\section{Sub-hypotheses about Perception, and Findings}
\label{appendix:modality-hypotheses}
We formulated and tested four hypotheses regarding how transformers perceive ARC grids, informed by the systematic results shown in Table~\ref{tab:modality-feature-accuracy}.

\paragraph{Sub-hypothesis 1: Tokenization Affects Spatial Encoding.}
Text modalities serialize 2D grids into 1D token sequences, requiring the transformer to reconstruct spatial relationships through attention. 
Image modalities preserve 2D structure natively and include explicit coordinate annotations, which help the model ground spatial references. 
However, images are tokenized by vision encoders into fixed-size patches, which can introduce \textit{aliasing} when fine details straddle patch boundaries or are under-sampled; 
this well-known sampling artifact can impair shift/translation precision at small scales \cite{zhang2019anti_aliasing}. 
For ARC, where single-cell cues matter, such aliasing can turn an otherwise correct localization into an off-by-one error—an effect we quantify later in our resolution sweeps.
The embedded coordinate labels bridge visual perception and symbolic references, reducing coordinate confusion errors observed in text-only modalities.

\begin{figure}[h]
    \centering
    \includegraphics[width=0.8\textwidth]{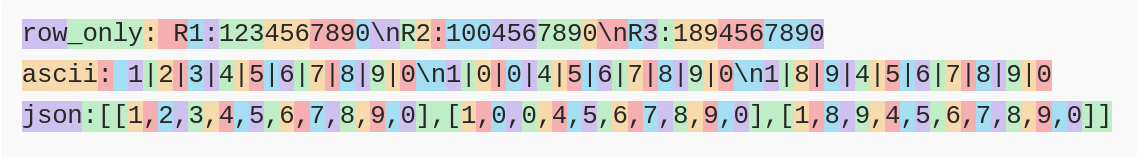}
    \caption{Visual comparison of tokenization for \texttt{row\_only}, \texttt{ascii}, and \texttt{json} formats using the same 3×10 grid example, generated using OpenAI's tokenizer \cite{openaitokenizer}. Each token is highlighted with a distinct color, showing how the same grid data is segmented differently across modalities. Note that different LLMs may tokenize the same input differently depending on their tokenizer implementation. The \texttt{row\_only} format often groups consecutive digits into multi-digit tokens (behavior is tokenizer-specific), \texttt{ascii} separates each digit and space, and \texttt{json} treats brackets and commas as distinct structural tokens.}
    \label{fig:tokenization_examples}
\end{figure}

To illustrate how different text modalities are tokenized, Figure~\ref{fig:tokenization_examples} visualizes how a sample 3×10 grid is segmented by OpenAI's tokenizer \cite{openaitokenizer}.
The key difference lies in how cell values are represented. In our compact \texttt{row\_only} representation, cell values (0--9) are concatenated without separators, allowing the tokenizer to group consecutive digits into multi-digit numeric tokens (e.g., 123, 100, 189). 
% Isn't it 
In contrast, \texttt{ascii} and \texttt{json} represent each cell as a single-digit value (0--9) with explicit separators. These differences have profound implications for how the model reconstructs 2D structure, which we explore in our experiments.

\textit{Hypothesis}: We hypothesized that different tokenization strategies would affect how well a model reconstructs 2D relationships. Specifically, formats with explicit separators (\texttt{ascii}, \texttt{json}) might provide clearer structural cues, while separator-free formats (\texttt{row\_only}) might enable more efficient reasoning by grouping digits into meaningful numeric tokens.
\textit{Findings}: Table~\ref{tab:modality-feature-accuracy} confirms that tokenization strategy has a significant impact, but its effectiveness is highly dependent on challenge characteristics. On challenge \texttt{13e47133}, which contains sparse but clearly delimited features (five isolated colored dots and a complex divider line), JSON achieved perfect accuracy (100\%), correctly identifying all five isolated colored dots with precise coordinates, while \texttt{row\_only} achieved 60.61\% by missing one pixel (Q11 instead of P11). ASCII achieved 87.32\%, outperforming \texttt{row\_only}, suggesting that explicit textual structure provides clearer cues for accurate coordinate reconstruction on structured tasks. On challenge \texttt{136b0064}, which contains eight distinct colored shapes arranged in four rows within a 15×15 grid, all text modalities achieve perfect accuracy (100\%): \texttt{row\_only}, \texttt{col\_only}, \texttt{ascii}, and \texttt{json} all correctly identify all eight shapes, the yellow vertical divider (H1-H15), and the grey dot (J1) with precise coordinates. This demonstrates that when features are well-structured and clearly delimited, different tokenization strategies can all achieve high accuracy, with explicit separators (\texttt{ascii}, \texttt{json}) and separator-free formats (\texttt{row\_only}) performing equally well. However, on challenge \texttt{0934a4d8}, which requires identifying a large block within a complex mosaic pattern with many scattered pixels, all text modalities struggled: JSON achieved the best performance at 30.00\%, while \texttt{row\_only} achieved only 5.00\% and ASCII achieved -11.67\%. This indicates that while tokenization helps on structured tasks, it does not overcome the fundamental bottleneck of 1D sequential processing for complex spatial patterns that lack simple structure. The trade-off is clear: no single tokenization strategy dominates across all challenge types, and the optimal format depends on how well the serialization direction and structural cues align with the challenge's spatial characteristics.

\paragraph{Sub-hypothesis 2: Text Modalities Induce a 1D Processing Bias.}
\textit{Hypothesis}: Given that text is processed sequentially, we predicted that row-wise formats (\texttt{row\_only}, \texttt{ascii}, \texttt{json}) would make horizontal features easier to identify than vertical features, which require attention across long token distances. Transposed formats (\texttt{col\_only}) should reverse this bias.
\textit{Findings}: Table~\ref{tab:modality-feature-accuracy} provides strong evidence for a 1D processing bias. On challenge \texttt{13e47133}, \texttt{row\_only} achieved 60.61\% final score, correctly identifying 4 of 5 isolated pixels including horizontal features (B2 and S2, which share rows with other pixels) but incorrectly identifying Q11 instead of P11 for a green pixel (off by 1 column), suggesting difficulty tracking vertical positions when processing horizontally. Conversely, \texttt{col\_only} achieved 70.00\% on \texttt{13e47133}, correctly identifying the vertical feature P11 that \texttt{row\_only} missed, but failed to identify the horizontal features B2 and S2 (identifying B1 instead of B2, S1 instead of S2, both off by 1 row). This bidirectional pattern---where each format excels at features aligned with its serialization direction and struggles with perpendicular features---provides compelling evidence for the 1D processing bias. The bias is most evident when comparing performance across challenges: on challenge \texttt{142ca369}, which contains four single dots in column K, \texttt{col\_only} achieves perfect accuracy (100\%) by correctly identifying all four pixels, while \texttt{row\_only} achieves 87.50\% by misidentifying K7 as L7 (off by 1 column), demonstrating that serialization direction determines which spatial relationships the model can reliably reconstruct. However, this advantage disappears on simpler, more structured tasks: on challenge \texttt{135a2760}, all text modalities achieve perfect accuracy (100\%), implying that the 1D processing bias is not a universal hindrance but emerges as a critical failure point when tasks require identifying sparse, non-local relationships that are misaligned with the serialization order.

\paragraph{Sub-hypothesis 3: Vision Model Perception Depends on Patch Alignment.}
\textit{Hypothesis}: For VLMs, which use patch-based tokenization, we hypothesized that perception accuracy depends on the relationship between cell size and patch size.
\textit{Findings}: Table~\ref{tab:modality-feature-accuracy} reveals \textit{non-linear resolution effects} that support this hypothesis, showing that more pixels are not always better. The detailed failure mode analysis in Appendix~\ref{appendix:perception-detail-result} demonstrates that most image modality errors stem from \textit{visual aliasing effects} introduced by patch-based tokenization. At low resolutions (14×14--17×17 pixels per cell), multiple grid cells pack into single vision patches, causing systematic hallucinations where single pixels are perceived as larger shapes. For example, on challenge \texttt{13e47133}, Image 14×14 hallucinated a ``Cyan 2x2 Square'' at B1:C2 when only B2 (teal) exists, and Image 17×17 described single pixels as ``Yellow Rectangle'' (2×1) and ``Green Rectangle'' (1×2). At high resolutions, single cells span multiple patches, leading to cross-boundary counting errors and coordinate misreadings when spreadsheet labels (e.g., ``A1'', ``P11'') straddle patch boundaries. Additionally, white coordinate annotations can alter patch embeddings, causing color misreadings: on challenge \texttt{13e47133}, Image 15×15 misread T1 (teal) as yellow, likely because white text overlay affected the patch's color embedding.

On challenge \texttt{13e47133}, which contains sparse isolated pixels, accuracy varies non-linearly with resolution: \texttt{image\_16x16} achieved only 45.73\% accuracy, while 24×24-1205 achieved 85.00\% and 24×24-1148 achieved 100\%, suggesting that the optimal resolution (approximately 1.5 patches per cell for Gemini's 16×16 patch size) prevents critical features from aligning precisely with patch boundaries. On challenge \texttt{0934a4d8}, which requires identifying a large contiguous block within a mosaic, boundary accuracy varies significantly: Image 15×15 and 24×24-1205 achieve the best performance (50.00\% and 45.00\%, respectively), while many other resolutions show significant errors or negative scores. Common errors include describing multiple blocks instead of one continuous region (e.g., Image 16×16 on \texttt{0934a4d8} split one teal block into three separate 4×4 blocks) or boundary misalignments (off by 1--5 columns/rows). On challenge \texttt{142ca369}, which contains scattered single pixels, most image modalities struggle severely (0--37.50\% accuracy), with only Image 17×17 achieving perfect accuracy (100\%), indicating that patch-based tokenization struggles with scattered pixels across large grids, though the exact failure pattern depends on resolution. This demonstrates that even with native 2D representation and explicit coordinates, VLM perception is susceptible to visual aliasing artifacts that depend critically on the relationship between cell size, patch size, and label placement.

\textit{Note on model-specificity}: Our experimental observations are based on Gemini 2.5 Pro. While the general principle that patch alignment affects perception should be generalizable to other VLMs, the exact optimal resolution and pixel-per-cell size are model-specific and depend on the vision encoder's patch size. For Gemini, based on public architecture information and our observations, we estimate a patch size of 16×16 pixels. Under this assumption, the 24×24 modality (24×24 pixels per cell) corresponds to approximately 1.5 patches per cell, or equivalently, a 2×2 cell region spans 3×3 patches. This non-integer mapping may help prevent critical features from aligning precisely with patch boundaries, potentially explaining why this resolution achieved better performance than other image modalities in our experiments. For 16×16, even though it matches the patch size of Gemini so avoid straddle problems, it compressed too much information into one patch, so we observed other failure modes like misreading cell color due to the annotation text. To generalize these findings to other models, researchers should either consult architecture documentation (when available) or empirically determine optimal resolutions through systematic perception benchmarks across different cell sizes, as the patch size varies across VLM architectures (e.g., Llama 3.2 Vision uses 14×14 patches, while GPT-4V's patch size is not publicly disclosed).

\paragraph{Sub-hypothesis 4: Multi-Modality Enables Cross-Validation.}
\textit{Hypothesis}: We predicted that combining text and image representations would improve accuracy by allowing the model to cross-reference complementary spatial information.
\textit{Findings}: Our perception experiments (Table~\ref{tab:modality-feature-accuracy}) tested individual modalities in isolation and revealed complementary strengths between text and image modalities, but did not directly test combined modalities in the perception task. However, our execution experiments (Section~\ref{subsec:multimodality-execution}) provide strong evidence for this hypothesis by demonstrating that multi-modal combinations consistently improve end-to-end reasoning accuracy. Figure~\ref{fig:cross_order_scatter_combined} shows that combining modalities (\texttt{row\_col\_json}, \texttt{row\_col\_json\_image}) achieves significantly higher median execution scores than single-modality encodings (\texttt{row\_only}, \texttt{col\_only}, \texttt{image\_only}, \texttt{json\_only}). The best-performing combination, \texttt{row\_col\_json\_image}, achieves a median similarity score of approximately 0.69—nearly double that of the worst-performing single modalities. This improvement stems from the model's ability to cross-validate spatial information across modalities: text modalities provide precise coordinate identification (as established in Sub-hypotheses 1--3), while image modalities confirm spatial relationships and shape patterns through native 2D representation. When combined, the model constructs a more accurate and complete internal representation, reducing the propagation of perception errors into reasoning failures. The key mechanism is that combining text with image modalities (\texttt{row\_col\_image}, \texttt{row\_col\_json\_image}) provides genuine complementarity by pairing sequential processing with native 2D representation, enabling true cross-modal validation. However, not all combinations provide complementarity: \texttt{row\_col} (combining only row-only and column-only text formats) performs worse than either format alone, suggesting that simply concatenating two 1D serializations does not enable effective cross-examination when both formats share the same fundamental limitation. Future work should conduct perception experiments with combined modalities to directly measure how multi-modal inputs affect feature identification accuracy, complementing the execution-based evidence presented here.

\section{Perception Failure Mode Analysis}
\label{appendix:perception-detail-result}

A detailed analysis reveals systematic failure modes that explain the accuracy differences. This section presents the ground truth features for each challenge, followed by modality-specific accuracy analysis.

\subsection{Challenge \texttt{13e47133}}

\paragraph{Ground Truth Features.}
This challenge contains the following critical features in a 20x20 grid:
\begin{itemize}[leftmargin=2em,noitemsep]
    \item \textbf{Five isolated colored dots}: A1 (orange), B2 (teal), T1 (teal), S2 (yellow), and P11 (green).
    \item \textbf{Complex multi-turn divider line} (red): The structure consists of four segments:
    \begin{itemize}[leftmargin=1.5em,noitemsep]
        \item Main vertical line: I1-I20
        \item Top horizontal line: I10-Q10 (or J10-Q10 if excluding left overlap, or J10-P10 if excluding both overlaps)
        \item Bottom horizontal line: I16-Q16 (or J16-Q16 if excluding left overlap, or J16-P16 if excluding both overlaps)
        \item Right vertical segment: Q10-Q16 (or Q11-Q15 if excluding both overlaps)
    \end{itemize}
    Note: Depending on whether to include or exclude overlapping pixels at junctions (I10, I16, Q10, Q16), multiple valid coordinate specifications exist.
    \item \textbf{Background color}: Blue (fills the entire canvas).
\end{itemize}

\paragraph{Scoring Guide.}
This challenge contains $N = 7$ features to track: five isolated colored dots (5 features each worth 15 points), one complex multi-turn divider line (1 feature worth 20 points), and one background color (1 binary feature worth 5 points). For each ground truth feature, we compute a similarity score starting at 100 and subtract penalties proportional to discrepancies. The penalty for each feature is $(points\_feature) \times |\mathrm{diff}|$, where $|\mathrm{diff}|$ is calculated using Equation~(1) in Section~\ref{subsec:perception-methodology}. For the background color (binary decision), the penalty is: 0 if correct, or 1 if wrong or no mention. The final accuracy score is $100 - \sum_{\mathrm{features}} \mathrm{penalty}$.

\paragraph{Accuracy Analysis.}

\subparagraph{Row-only.}
\textbf{Five isolated colored dots}: Correctly identified 4 of 5 pixels (A1 orange, T1 teal, B2 teal, S2 yellow). Found Q11 instead of P11 (green), off by 1 column.
\textbf{Complex multi-turn divider line}: Claimed ``Main Vertical Line: J1-J20; Top Horizontal Line: J10-R10; Bottom Horizontal Line: J16-R16; Right Vertical Segment: R11-R15''. All four boundaries wrong: main vertical at J instead of I (off by 1 column), horizontal lines J-R instead of I-Q (off by 1 column on each end), right vertical at R instead of Q (off by 1 column).
\textbf{Background color}: Correctly identified blue (1) as the predominant background color.

\subparagraph{Column-only.}
\textbf{Five isolated colored dots}: Correctly identified 3 of 5 pixels (A1 orange, T1 teal, P11 green). Found B1 instead of B2 (teal), and S1 instead of S2 (yellow), both off by 1 row.
\textbf{Complex multi-turn divider line}: Claimed ``Full Vertical Line: I1-I20; Top Bar: I10-Q10; Bottom Bar: I16-Q16; Right Bar: Q10-Q16''. Correctly identified main vertical line (I1-I20), top horizontal line (I10-Q10), and bottom horizontal line (I16-Q16). Failed on right vertical segment (Q10-Q16 instead of Q11-Q15).
\textbf{Background color}: Correctly identified blue (1) as the predominant background color.

\subparagraph{ASCII.}
\textbf{Five isolated colored dots}: Achieved perfect coordinate identification for all five pixels (A1 orange, B2 teal, T1 teal, S2 yellow, P11 green).
\textbf{Complex multi-turn divider line}: Claimed ``Main Vertical Stem: I1-I20; Top Horizontal Bar: J9-J17 (row J); Bottom Horizontal Bar: P9-P17 (row P); Right Vertical Bar: Q11-Q15''. Correctly identified main vertical line (I1-I20) and right vertical segment (Q11-Q15). Failed on top horizontal line (described as row J from J9-J17, but should be row 10 from I10-Q10 or J10-Q10, confused row with column) and bottom horizontal line (described as row P from P9-P17, but should be row 16 from I16-Q16 or J16-Q16, confused row with column).
\textbf{Background color}: Correctly identified blue (1) as the predominant background color.

\subparagraph{JSON.}
\textbf{Five isolated colored dots}: Achieved perfect coordinate identification for all five pixels (A1 orange, B2 teal, T1 teal, S2 yellow, P11 green).
\textbf{Complex multi-turn divider line}: Claimed ``Vertical Spine: I1-I20; Top Border: I10-Q10; Bottom Border: I16-Q16; Right Border: Q11-Q15''. Correctly identified all four boundaries: main vertical (I1-I20), top horizontal (I10-Q10), bottom horizontal (I16-Q16), right vertical (Q11-Q15).
\textbf{Background color}: Correctly identified blue (1) as the background color filling all other cells.

\subparagraph{Image 14×14.}
\textbf{Five isolated colored dots}: Hallucinated ``Cyan 2x2 Square'' at B1:C2 (ground truth: A1 orange, B2 teal - hallucinated extra pixels at B1, C1, C2) and ``Yellow L-Shape'' at S1+T1+S2 (ground truth: T1 teal, S2 yellow - hallucinated extra pixel at S1). Correctly identified P11 (green). The hallucinations are evidence of visual aliasing effect.
\textbf{Complex multi-turn divider line}: Claimed ``Vertical Line: I1-I20; Hollow Rectangle: Top edge I10-R10, Right edge R10-R16, Bottom edge R16-I16, Left edge I16-I10''. Correctly identified main vertical line (I1-I20). Failed on all rectangle boundaries: top horizontal (I10-R10 instead of I10-Q10, off by 1 column on right), bottom horizontal (R16-I16 instead of I16-Q16, off by 1 column on right), right vertical (R10-R16 instead of Q11-Q15, off by 1 column and wrong row range).
\textbf{Background color}: Correctly identified dark blue (1) as the background color.

\subparagraph{Image 15×15.}
\textbf{Five isolated colored dots}: Correctly identified A1 (orange), P11 (green). Found B1 instead of B2 (teal, off by 1 row). Found only T1 (yellow) when ground truth has T1 (teal) and S2 (yellow) - missed T1 teal entirely, color confusion on T1. Missing S2 (yellow).
\textbf{Complex multi-turn divider line}: Claimed ``Vertical Stem: I2-I9; Rectangular Frame: Top Edge I10-Q10, Bottom Edge I16-Q16, Left Edge I10-I16, Right Edge Q10-Q16''. Correctly identified top horizontal (I10-Q10) and bottom horizontal (I16-Q16). Failed on main vertical (I2-I9 instead of I1-I20, missing rows 1 and 10-20) and right vertical (Q10-Q16 instead of Q11-Q15, included overlap points).
\textbf{Background color}: Correctly identified blue (1) as the uniform background color.

\subparagraph{Image 16×16.}
\textbf{Five isolated colored dots}: Correctly identified A1 (orange). Found B1 instead of B2 (teal, off by 1 row), S1 instead of S2 (yellow, off by 1 row), Q11 instead of P11 (green, off by 1 column). Missing T1 (teal).
\textbf{Complex multi-turn divider line}: Described fragmented segments (``Upper Vertical Stem: J2-J9; Top Horizontal: J10-Q10; Left Vertical: I10-I15; Right Vertical: Q12-Q15; Bottom Horizontal: I16-Q16; Lower Vertical Stem: I17-I19'') instead of coherent structure. Correctly identified bottom horizontal (I16-Q16). Failed on main vertical (described as split segments J2-J9, I10-I15, I17-I19 in different columns instead of continuous I1-I20), top horizontal (J10-Q10 instead of I10-Q10, off by 1 column on left), right vertical (Q12-Q15 instead of Q11-Q15, off by 1 row on top).
\textbf{Background color}: Correctly identified light blue (1) as the uniform background color.

\subparagraph{Image 17×17.}
\textbf{Five isolated colored dots}: Correctly identified A1 (orange), B2 (teal, described as Light Blue/Cyan but correct location). Hallucinated S2 (yellow) as ``Yellow Rectangle'' at R2+S2 (2x1 rectangle, added non-existent R2), P11 (green) as ``Green Rectangle'' at P11+P12 (1x2 rectangle, added non-existent P12). Missing T1 (teal). The hallucinations are evidence of visual aliasing effect.
\textbf{Complex multi-turn divider line}: Claimed ``Column I: I1-I16; Row 10: J10-Q10; Row 16: J16-Q16; Column Q: Q11-Q15''. Correctly identified right vertical (Q11-Q15). Failed on main vertical (I1-I16 instead of I1-I20, missing rows 17-20), top horizontal (J10-Q10 instead of I10-Q10, off by 1 column on left), bottom horizontal (J16-Q16 instead of I16-Q16, off by 1 column on left).
\textbf{Background color}: Correctly identified blue (1) as the background color.

\subparagraph{24×24-1148.}
\textbf{Five isolated colored dots}: Correctly identified A1 (orange), T1 (teal). Hallucinated B2 (teal) into ``Light Blue Rectangle (1x2)'' at A2:B2 (added non-existent A2), S2 (yellow) into ``Yellow Square (2x2)'' at S2:T3 (added non-existent T2, S3, T3), P11 (green) into ``Green Square (2x2)'' at P11:Q12 (added non-existent Q11, P12, Q12). The hallucinations are evidence of visual aliasing effect.
\textbf{Complex multi-turn divider line}: Claimed ``Vertical Line: I1-I20; Rectangular Loop: Top edge I10-Q10, Right edge Q10-Q16, Bottom edge I16-Q16, Left edge I10-I16''. Correctly identified main vertical (I1-I20), top horizontal (I10-Q10), bottom horizontal (I16-Q16). Failed on right vertical (Q10-Q16 instead of Q11-Q15, included overlap points).
\textbf{Background color}: Correctly identified blue (1) as the uniform background color.

\subparagraph{24×24-1205.}
\textbf{Five isolated colored dots}: Correctly identified A1 (orange), B2 (teal, described as light blue/cyan but correct location), S2 (yellow), P11 (green). Missing T1 (teal).
\textbf{Complex multi-turn divider line}: Claimed ``Hollow Rectangle with Vertical Line/Tail: Vertical Line I1-I20, Top Horizontal Line J10-Q10, Bottom Horizontal Line J16-Q16, Right Vertical Line Q11-Q15''. Correctly identified main vertical (I1-I20) and right vertical (Q11-Q15). Failed on top horizontal (J10-Q10 instead of I10-Q10, off by 1 column on left) and bottom horizontal (J16-Q16 instead of I16-Q16, off by 1 column on left).
\textbf{Background color}: Correctly identified blue (1) as the solid background color.

\subparagraph{Image 768×768.}
\textbf{Five isolated colored dots}: Correctly identified A1 (orange), B2 (teal), P11 (green). Hallucinated combining T1 (teal) and S2 (yellow) into ``Yellow Rectangle (1x2)'' at S1 and T1 (wrong row for both pixels, merged two separate colored pixels with different colors into one yellow shape).
\textbf{Complex multi-turn divider line}: Claimed ``First Vertical Segment: I1-I20; First Horizontal Segment: J10-Q10; Second Vertical Segment: Q11-Q16; Second Horizontal Segment: J16-P16''. Correctly identified main vertical (I1-I20). Failed on top horizontal (J10-Q10 instead of I10-Q10, off by 1 column on left), bottom horizontal (J16-P16 instead of I16-Q16, off by 1 column on each end), right vertical (Q11-Q16 instead of Q11-Q15, off by 1 row on bottom).
\textbf{Background color}: Correctly identified medium blue (1) as the predominant background color.

\paragraph{Final Scores.}
\begin{table}[h]
\centering
\small
\begin{tabular}{lr}
\toprule
Modality & Final Score \\
\midrule
Row-only & 60.61 \\
Column-only & 70.00 \\
ASCII & 87.32 \\
JSON & 100.00 \\
Image 14×14 & 49.15 \\
Image 15×15 & 52.56 \\
Image 16×16 & 45.73 \\
Image 17×17 & 68.05 \\
24×24-1148 & 100.00 \\
24×24-1205 & 85.00 \\
Image 768×768 & 70.00 \\
\bottomrule
\end{tabular}
\end{table}

\subsection{Challenge \texttt{135a2760}}

\paragraph{Ground Truth Features.}
This challenge contains the following critical features in a 13x5 grid:
\begin{itemize}[leftmargin=2em,noitemsep]
    \item \textbf{Nested green rectangular frames}: The outer frame spans A1:M1 (top edge), A5:M5 (bottom edge), A1:A5 (left edge), and M1:M5 (right edge), forming a rectangle A1:M5 containing 27 pixels total.
    \item \textbf{Nested red rectangular frames}: The outer frame spans B2:L2 (top edge), B4:L4 (bottom edge), B2:B4 (left edge), and L2:L4 (right edge), forming a rectangle B2:L4.
    \item \textbf{Blue dots}: Four isolated blue pixels at coordinates C3, E3, G3, and K3.
\end{itemize}

\paragraph{Scoring Guide.}
This challenge contains $N = 6$ features to track: nested green rectangular frames (1 feature worth 10 points), nested red rectangular frames (1 feature worth 10 points), and four blue dots (4 features each worth 20 points). For each ground truth feature, we compute a similarity score starting at 100 and subtract penalties proportional to discrepancies. The penalty for each feature is $(points\_feature) \times |\mathrm{diff}|$, where $|\mathrm{diff}|$ is calculated using Equation~(1) in Section~\ref{subsec:perception-methodology}. The final accuracy score is $100 - \sum_{\mathrm{features}} \mathrm{penalty}$.

\paragraph{Accuracy Analysis.}

\subparagraph{Row-only.}
\textbf{Nested green rectangular frames}: Correctly identified all four edges: top edge A1-M1, bottom edge A5-M5, left edge A1-A5, right edge M1-M5.
\textbf{Nested red rectangular frames}: Correctly identified all four edges: top edge B2-L2, bottom edge B4-L4, left edge B2-B4, right edge L2-L4.
\textbf{Blue dots}: Claimed five blue pixels at C3, E3, G3, I3, K3. Found extra dot at I3 (which is green, not blue). Correctly identified C3, E3, G3, K3 but included incorrect I3.

\subparagraph{Column-only.}
\textbf{Nested green rectangular frames}: Correctly identified all four edges: top edge A1-M1, bottom edge A5-M5, left edge A1-A5, right edge M1-M5.
\textbf{Nested red rectangular frames}: Correctly identified all four edges: top edge B2-L2, bottom edge B4-L4, left edge B2-B4, right edge L2-L4.
\textbf{Blue dots}: Correctly identified all four blue pixels at C3, E3, G3, and K3.

\subparagraph{ASCII.}
\textbf{Nested green rectangular frames}: Correctly identified all four edges: top edge A1-M1, bottom edge A5-M5, left edge A1-A5, right edge M1-M5.
\textbf{Nested red rectangular frames}: Correctly identified all four edges: top edge B2-L2, bottom edge B4-L4, left edge B2-B4, right edge L2-L4.
\textbf{Blue dots}: Correctly identified all four blue pixels at C3, E3, G3, and K3.

\subparagraph{JSON.}
\textbf{Nested green rectangular frames}: Correctly identified all four edges: top edge A1-M1, bottom edge A5-M5, left edge A1-A5, right edge M1-M5.
\textbf{Nested red rectangular frames}: Correctly identified all four edges: top edge B2-L2, bottom edge B4-L4, left edge B2-B4, right edge L2-L4.
\textbf{Blue dots}: Correctly identified all four blue pixels at C3, E3, G3, and K3.

\subparagraph{Image 14×14.}
\textbf{Nested green rectangular frames}: Correctly identified all four edges as background/frame: top edge A1-M1, bottom edge A5-M5, left edge A1-A5, right edge M1-M5.
\textbf{Nested red rectangular frames}: Correctly identified structure: top bar B2-L2, bottom bar B4-L4, left connector B3, right connector L3.
\textbf{Blue dots}: Hallucinated three blue blocks instead of four isolated dots: Block 1 (3x1 rectangle) at C3-D3-E3 (C3 and E3 are correct blue pixels, but D3 is green, not blue), Block 2 (2x1 rectangle) at G3-H3 (G3 is correct blue pixel, but H3 is green, not blue), Block 3 (1x1 square) at K3 (correct). The hallucinations are evidence of visual aliasing effect where adjacent pixels of different colors were merged into rectangular blocks.

\subparagraph{Image 15×15.}
\textbf{Nested green rectangular frames}: Correctly identified all four edges: top edge A1-M1, bottom edge A5-M5, left edge A1-A5, right edge M1-M5.
\textbf{Nested red rectangular frames}: Correctly identified rectangle B2-L4 (described as 3x11 solid red rectangle, though it's actually a hollow frame with top bar B2-L2, bottom bar B4-L4).
\textbf{Blue dots}: Correctly identified only 3 of 4 blue pixels: C3, G3, K3. Missing E3.

\subparagraph{Image 16×16.}
\textbf{Nested green rectangular frames}: Correctly identified all four edges: top edge A1-M1, bottom edge A5-M5, left edge A1-A5, right edge M1-M5.
\textbf{Nested red rectangular frames}: Correctly identified rectangle B2-L4 (described as solid 11x3 rectangle, though it's actually a hollow frame with top bar B2-L2, bottom bar B4-L4).
\textbf{Blue dots}: Claimed five blue pixels at C3, E3, G3, I3, K3. Found extra dot at I3 (which is green, not blue). Correctly identified C3, E3, G3, K3 but included incorrect I3.

\subparagraph{Image 17×17.}
\textbf{Nested green rectangular frames}: Correctly identified all four edges: top edge A1-M1, bottom edge A5-M5, left edge A1-A5, right edge M1-M5.
\textbf{Nested red rectangular frames}: Correctly identified rectangle B2-L4 (described as 3x11 inner rectangle, though it's actually a hollow frame with top bar B2-L2, bottom bar B4-L4).
\textbf{Blue dots}: Claimed five blue pixels at C3, E3, G3, I3, K3. Found extra dot at I3 (which is green, not blue). Correctly identified C3, E3, G3, K3 but included incorrect I3.

\subparagraph{24×24-1148.}
\textbf{Nested green rectangular frames}: Correctly identified all four edges: top edge A1-M1, bottom edge A5-M5, left edge A1-A5, right edge M1-M5.
\textbf{Nested red rectangular frames}: Correctly identified rectangle B2-L4 with top bar B2-L2, bottom bar B4-L4, vertical side-bars at B3 and L3.
\textbf{Blue dots}: Correctly identified all four blue pixels at C3, E3, G3, and K3.

\subparagraph{24×24-1205.}
\textbf{Nested green rectangular frames}: Correctly identified all four edges: top edge A1-M1, bottom edge A5-M5, left edge A1-A5, right edge M1-M5.
\textbf{Nested red rectangular frames}: Correctly identified rectangle B2-L4 (described as 3x11 large solid red rectangle, though it's actually a hollow frame with top bar B2-L2, bottom bar B4-L4).
\textbf{Blue dots}: Correctly identified all four blue pixels at C3, E3, G3, and K3.

\subparagraph{Image 768×768.}
\textbf{Nested green rectangular frames}: Correctly identified all four edges: top edge A1-M1, bottom edge A5-M5, left edge A1-A5, right edge M1-M5.
\textbf{Nested red rectangular frames}: Correctly identified rectangle B2-L4 with top part B2-L2, bottom part B4-L4, vertical side-bars B3 and L3.
\textbf{Blue dots}: Correctly identified all four blue pixels at C3, E3, G3, and K3.

\paragraph{Final Scores.}
\begin{table}[h]
\centering
\small
\begin{tabular}{lr}
\toprule
Modality & Final Score \\
\midrule
Row-only & 100.00 \\
Column-only & 100.00 \\
ASCII & 100.00 \\
JSON & 100.00 \\
Image 14×14 & 100.00 \\
Image 15×15 & 76.25 \\
Image 16×16 & 96.25 \\
Image 17×17 & 96.25 \\
24×24-1148 & 100.00 \\
24×24-1205 & 96.25 \\
Image 768×768 & 100.00 \\
\bottomrule
\end{tabular}
\end{table}

\subsection{Challenge \texttt{142ca369}}

\paragraph{Ground Truth Features.}
This challenge contains the following critical features in a 20x20 grid:
\begin{itemize}[leftmargin=2em,noitemsep]
    \item \textbf{Four L-shaped objects}:
    \begin{itemize}[leftmargin=1.5em,noitemsep]
        \item Green L-shape: A9, B9, B8 (3 pixels)
        \item Blue L-shape: C7, D7, D6 (3 pixels)
        \item Red L-shape: E5, F5, F4 (3 pixels)
        \item Grey L-shape: G3, H3, H2 (3 pixels)
    \end{itemize}
    \item \textbf{Single dots}: Four isolated pixels at coordinates K7, K11, K15, and K19.
\end{itemize}

\paragraph{Scoring Guide.}
This challenge contains $N = 8$ features to track: four L-shaped objects (4 features) and four single dots (4 features), each worth 12.5 points. For each ground truth feature, we compute a similarity score starting at 100 and subtract penalties proportional to discrepancies. The penalty for each feature is $(12.5) \times |\mathrm{diff}|$, where $|\mathrm{diff}|$ is calculated using Equation~(1) in Section~\ref{subsec:perception-methodology}. The final accuracy score is $100 - \sum_{\mathrm{features}} \mathrm{penalty}$.

\paragraph{Accuracy Analysis.}

\subparagraph{Row-only.}
\textbf{Four L-shaped objects}: Correctly identified all four L-shapes with correct coordinates: Gray L-shape (H2, G3, H3), Red L-shape (F4, E5, F5), Blue L-shape (D6, C7, D7), Green L-shape (B8, A9, B9).
\textbf{Single dots}: Correctly identified 3 of 4 pixels (K11 red, K15 blue, K19 green). Found L7 instead of K7 (gray), off by 1 column.

\subparagraph{Column-only.}
\textbf{Four L-shaped objects}: Correctly identified all four L-shapes with correct coordinates: Gray (H2, G3, H3), Red (F4, E5, F5), Blue (D6, C7, D7), Green (B8, A9, B9).
\textbf{Single dots}: Correctly identified all four pixels at K7, K11, K15, and K19.

\subparagraph{ASCII.}
\textbf{Four L-shaped objects}: Correctly identified Red, Blue, and Green L-shapes. Failed on Gray L-shape: claimed (G2, G3, H3) but ground truth is (H2, G3, H3) - missing H2 and incorrectly included G2 (off by 1 row).
\textbf{Single dots}: Correctly identified all four pixels at K7, K11, K15, and K19.

\subparagraph{JSON.}
\textbf{Four L-shaped objects}: Correctly identified all four L-shapes with correct coordinates: Gray (H2, G3, H3), Red (F4, E5, F5), Blue (D6, C7, D7), Green (B8, A9, B9).
\textbf{Single dots}: Correctly identified all four pixels at K7, K11, K15, and K19.

\subparagraph{Image 14×14.}
\textbf{Four L-shaped objects}: Failed to recognize the four 3-cell L-shapes. Instead perceived them as eight separate 2-cell vertical blocks labeled A-H, forming a diagonal chain. Completely misidentified the structure despite the ground truth pixels being present.
\textbf{Single dots}: Claimed K5 instead of K7 (gray, off by 2 rows), correctly identified K11 (red) and K15 (blue). Missing K19 (green).

\subparagraph{Image 15×15.}
\textbf{Four L-shaped objects}: Failed to recognize the four 3-cell L-shapes. Instead perceived them as four 4-cell vertical bars (each 4x1). Completely misidentified the structure and hallucinated additional pixels A8, A10, A11, M21 not in ground truth.
\textbf{Single dots}: Claimed J9 instead of K7 (gray, off by 1 column and 2 rows). Missing K11, K15, K19.

\subparagraph{Image 16×16.}
\textbf{Four L-shaped objects}: Failed to recognize the four 3-cell L-shapes. Instead perceived them as four 3-cell vertical blocks (each 3x1). Completely misidentified the structure and hallucinated additional pixels B3, B4, B5 not in ground truth.
\textbf{Single dots}: Correctly identified K7 (gray), K11 (red), and K15 (blue). Missing K19 (green).

\subparagraph{Image 17×17.}
\textbf{Four L-shaped objects}: Correctly identified all four L-shapes with accurate coordinates: Gray (H2, G3, H3), Red (F4, E5, F5), Blue (D6, C7, D7), Green (B8, A9, B9). However, also included hallucinated extra objects including a blue 2-cell domino (D6, E6) and hallucinated additional pixel B10 not in ground truth.
\textbf{Single dots}: Correctly identified all four pixels at K7, K11, K15, and K19.

\subparagraph{24×24-1148.}
\textbf{Four L-shaped objects}: Failed to recognize the four 3-cell L-shapes. Instead perceived the green L-shape as two separate 3x2 rectangles (A3:C4 and D3:F4), and perceived the Blue, Red, and Gray L-shapes as 5-cell L-shapes (each with a 3-cell horizontal bar and 2-cell vertical stem). Completely misidentified the structure and hallucinated massive additional pixels A3-B4, D3-F4, and many more not in ground truth.
\textbf{Single dots}: Hallucinated all single pixels into 2x2 squares: Gray square at M7-N8 (should be single pixel K7), Red square at M11-N12 (should be K11), Blue square at M15-N16 (should be K15), Green square at M19-N20 (should be K19). The hallucinations are evidence of visual aliasing effect.

\subparagraph{24×24-1205.}
\textbf{Four L-shaped objects}: Failed to recognize the four 3-cell L-shapes. Instead perceived each L-shape as a complex 6-cell structure consisting of a 2x2 square attached to a 1x4 vertical rectangle. Completely misidentified the structure and hallucinated massive additional pixels including A5:B6, C5:C8, D4:F7, G3:I6, J2:L5, M21-N22 not in ground truth.
\textbf{Single dots}: Hallucinated all single pixels into 2x2 squares: Gray square at M9:N10 (should be single pixel K7), Red square at M13:N14 (should be K11), Blue square at M17:N18 (should be K15), Green square at M21:N22 (should be K19). The hallucinations are evidence of visual aliasing effect.

\subparagraph{Image 768×768.}
\textbf{Four L-shaped objects}: Failed to recognize the four 3-cell L-shapes. Instead perceived each L-shape as a 4-cell L-shape consisting of one horizontal cell and three vertical cells. Hallucinated additional pixel in each L-shape: Green L described as (A5, B5, B6, B7) instead of (B8, A9, B9), Blue L as (C4, D4, D5, D6) instead of (D6, C7, D7), Red L as (E3, F3, F4, F5) instead of (F4, E5, F5), Gray L as (G2, H2, H3, H4) instead of (H2, G3, H3). Completely misidentified the structure.
\textbf{Single dots}: Correctly identified all four pixels at K7, K11, K15, and K19.

\paragraph{Final Scores.}
\begin{table}[h]
\centering
\small
\begin{tabular}{lr}
\toprule
Modality & Final Score \\
\midrule
Row-only & 87.50 \\
Column-only & 100.00 \\
ASCII & 91.67 \\
JSON & 100.00 \\
Image 14×14 & 25.00 \\
Image 15×15 & 0.00 \\
Image 16×16 & 37.50 \\
Image 17×17 & 100.00 \\
24×24-1148 & 0.00 \\
24×24-1205 & 0.00 \\
Image 768×768 & 25.00 \\
\bottomrule
\end{tabular}
\end{table}

\subsection{Challenge \texttt{136b0064}}

\paragraph{Ground Truth Features.}
This challenge contains the following critical features in a 15x15 grid:
\begin{itemize}[leftmargin=2em,noitemsep]
    \item \textbf{Eight distinct colored shapes arranged in four rows}:
    \begin{itemize}[leftmargin=1.5em,noitemsep]
        \item \textbf{Row 1}: Red U-shaped (A1, C1, A2, C2, A3, B3, C3 - 7 pixels), Pink Y-shaped (E1, G1, F2, F3 - 4 pixels)
        \item \textbf{Row 2}: Blue Q-shaped (A5, B5, A6, C6, B7 - 5 pixels), Green M-shaped (E5, F5, G5, F6, E7, G7 - 6 pixels)
        \item \textbf{Row 3}: Blue Q-shaped (A9, B9, A10, C10, B11 - 5 pixels), Pink Y-shaped (E9, G9, F10, F11 - 4 pixels)
        \item \textbf{Row 4}: Pink Y-shaped (A13, C13, B14, B15 - 4 pixels), Blue Q-shaped (E13, F13, E14, G14, F15 - 5 pixels)
    \end{itemize}
    \item \textbf{Yellow vertical divider}: The entire column H from H1 to H15 (15 pixels).
    \item \textbf{Grey dot}: Single isolated pixel at J1.
\end{itemize}

\paragraph{Scoring Guide.}
This challenge contains $N = 10$ features to track: eight distinct colored shapes (8 features each worth 10 points), one yellow vertical divider (1 feature worth 10 points), and one grey dot (1 feature worth 10 points). For each ground truth feature, we compute a similarity score starting at 100 and subtract penalties proportional to discrepancies. The penalty for each feature is $(points\_feature) \times |\mathrm{diff}|$, where $|\mathrm{diff}|$ is calculated using Equation~(1) in Section~\ref{subsec:perception-methodology}. The final accuracy score is $100 - \sum_{\mathrm{features}} \mathrm{penalty}$.

\paragraph{Accuracy Analysis.}

\subparagraph{Row-only.}
\textbf{Eight distinct colored shapes}: Correctly identified all eight shapes with accurate coordinates: Red U-shape at A1:C3 (A1, C1, A2, C2, A3, B3, C3), Pink Y-shape at E1:G3 (E1, G1, F2, F3), Blue Q-shape at A5:C7 (A5, B5, A6, C6, B7), Green M-shape at E5:G7 (E5, F5, G5, F6, E7, G7), Blue Q-shape at A9:C11 (A9, B9, A10, C10, B11), Pink Y-shape at E9:G11 (E9, G9, F10, F11), Pink Y-shape at A13:C15 (A13, C13, B14, B15), Blue Q-shape at E13:G15 (E13, F13, E14, G14, F15).
\textbf{Yellow vertical divider}: Correctly identified H1-H15.
\textbf{Grey dot}: Correctly identified at J1.

\subparagraph{Column-only.}
\textbf{Eight distinct colored shapes}: Correctly identified all eight shapes with accurate coordinates: Red U-shape at A1:C3 (A1, A2, A3, B3, C1, C2, C3), Pink shapes at E1:G3 (E1, G1, F2, F3), E9:G11 (E9, G9, F10, F11), and A13:C15 (A13, C13, B14, B15), Blue shapes at A5:C7 (A5, B5, A6, C6, B7), A9:C11 (A9, B9, A10, C10, B11), and E13:G15 (E13, F13, E14, G14, F15), Green M-shape at E5:G7 (E5, F5, G5, F6, E7, G7).
\textbf{Yellow vertical divider}: Correctly identified H1-H15.
\textbf{Grey dot}: Correctly identified at J1.

\subparagraph{ASCII.}
\textbf{Eight distinct colored shapes}: Correctly identified all eight shapes with accurate coordinates: Red U-shape at A1:C3 (A1, C1, A2, C2, A3, B3, C3), Pink ``Staple'' shapes (described as upside-down Pi) at E1:G3 (E1, G1, F2, F3), E9:G11 (E9, G9, F10, F11), and A13:C15 (A13, C13, B14, B15), Blue ``Rotated C'' shapes at A5:C7 (A5, B5, A6, C6, B7), A9:C11 (A9, B9, A10, C10, B11), and E13:G15 (E13, F13, E14, G14, F15), Green hollow square at E5:G7 (E5, F5, G5, F6, E7, G7).
\textbf{Yellow vertical divider}: Correctly identified H1-H15.
\textbf{Grey dot}: Correctly identified at J1.

\subparagraph{JSON.}
\textbf{Eight distinct colored shapes}: Correctly identified all eight shapes with accurate coordinates, described as stylized pixel-art letters: Red 'R'-like at A1:C3 (A1, C1, A2, C2, A3, B3, C3), Pink 'M'-like shapes at E1:G3 (E1, G1, F2, F3), E9:G11 (E9, G9, F10, F11), and A13:C15 (A13, C13, B14, B15), Blue 'B'-like shapes at A5:C7 (A5, B5, A6, C6, B7), A9:C11 (A9, B9, A10, C10, B11), and E13:G15 (E13, F13, E14, G14, F15), Green 'G'-like at E5:G7 (E5, F5, G5, F6, E7, G7).
\textbf{Yellow vertical divider}: Correctly identified H1-H15.
\textbf{Grey dot}: Correctly identified at J1.

\subparagraph{Image 14×14.}
\textbf{Eight distinct colored shapes}: Correctly identified all eight shapes using 0-indexed rows: Red U-shape at A0-C2 (A0, A1, A2, B2, C0, C1, C2), Pink upside-down T at E0-G1 (E0, F0, G0, F1), Blue shapes at A4-C6 (A4, A5, B4, C4, C5, B6), A8-C11 (A8, A9, A10, B8, B11, C8, C9, C10), and E13-G15 (E13, F13, G13, F14, G14, F15), Green arch at E4-G6 (E4, F4, G4, F5, E6, G6), Pink shapes at E8-G11 (E8, F8, G8, F9, F10, F11) and A13-C15 (A13, B14, C13, B15). All coordinates match ground truth when adjusted for 0-indexing.
\textbf{Yellow vertical divider}: Correctly identified H0-H15 (using 0-indexed rows).
\textbf{Grey dot}: Hallucinated as ``Gray Domino Shape'' 1x2 horizontal rectangle at J0 and K0 (added extra pixel K0 that doesn't exist, should be single pixel at J0 in 0-indexed system).

\subparagraph{Image 15×15.}
\textbf{Eight distinct colored shapes}: Correctly identified all eight shapes with accurate coordinates: Red U-shape at A1-C3 (A1, A2, A3, B3, C1, C2, C3), Pink inverted T at E1-G3 (E1, F1, G1, F2, F3), Blue C-shape at A5-C7 (A5, B5, A6, C6, B7), Green asymmetric shape at E5-G7 (E5, F5, G5, F6, G6, E7, G7 - note: described as 7 cells but matches ground truth 6 cells E5, F5, G5, F6, E7, G7), Blue irregular shape at A9-C11 (A9, B9, A10, C10, B11), Pink Y-shape at E9-G11 (E9, G9, F10, F11), Pink T-shape at A13-C15 (A13, B13, C13, B14, B15 - described as upright T with top bar A13-B13-C13, but matches ground truth A13, C13, B14, B15), Blue/multi-colored shape at E13-G15 (E13, F13, E14, G14, F15).
\textbf{Yellow vertical divider}: Correctly identified H1-H15.
\textbf{Grey dot}: Correctly identified at J1.

\subparagraph{Image 16×16.}
\textbf{Eight distinct colored shapes}: Correctly identified all eight shapes with accurate coordinates: Red U-shape at A1-C3 (A1, A2, A3, B3, C1, C2, C3), Pink inverted T at E1-G3 (E1, F1, G1, F2, F3), Blue pinwheel shapes at A5-C7 (A5, B5, A6, C6, B7) and A9-C11 (A9, B9, A10, C10, B11), Green incomplete square at E5-G7 (E5, F5, G5, E6, G6, E7, G7 - described as 8-cell square missing F7, but matches ground truth 6 cells E5, F5, G5, F6, E7, G7), Pink Y-shape at E9-G11 (E9, G9, F10, F11), Pink inverted Y at A13-C15 (A13, C13, B14, B15), Blue asymmetric shape at E13-G15 (E13, F13, E14, G14, F15).
\textbf{Yellow vertical divider}: Correctly identified H1-H15.
\textbf{Grey dot}: Correctly identified at J1.

\subparagraph{Image 17×17.}
\textbf{Eight distinct colored shapes}: Correctly identified all eight shapes with accurate coordinates: Red U-shape at A1-C3 (A1, A2, A3, B3, C1, C2, C3), Pink T-shape at E1-G3 (E1, F1, G1, F2, F3), Blue C-shapes at A5-C7 (A5, A6, B5, B7, C6) and A9-C11 (A9, A10, B9, B11, C10), Green symmetrical shape at E5-G7 (E5, F5, G5, F6, E7, G7), Pink Y-shape at E9-G11 (E9, G9, F10, F11), Pink inverted T at A13-C15 (A13, C13, B14, B15), Blue F-like shape at E13-G15 (E13, F13, G13, F14, F15 - described as horizontal bar with stem, but coordinates match ground truth E13, F13, E14, G14, F15).
\textbf{Yellow vertical divider}: Correctly identified H1-H15.
\textbf{Grey dot}: Correctly identified at J1.

\subparagraph{24×24-1148.}
\textbf{Eight distinct colored shapes}: Correctly identified all eight shapes with mostly accurate coordinates: Red U-shape at A1-C3 (A1, A2, A3, B3, C3, C2, C1), Pink T-shape at E1-G3 (E1, F1, G1, F2, F3), Blue plus-shapes at A5-C7 (B5, A6, B6, C6, B7) and A9-C11 (B9, A10, B10, C10, B11), Green T-shape at E5-G7 (E5, F5, G5, F6, F7 - described as having stem F6, F7, but ground truth is E5, F5, G5, F6, E7, G7), Pink T-shape at E9-G11 (E9, F9, G9, F10, F11 - described with coordinates E9, G9, F10, F11 but claimed F9 instead), Pink L-shape at A13-C15 (A13, B13, B14, B15 - matches ground truth A13, C13, B14, B15 except claimed B13 instead of C13), Blue modified T at E13-G15 (F13, E14, F14, G14, F15 - matches ground truth E13, F13, E14, G14, F15 except missing E13).
\textbf{Yellow vertical divider}: Correctly identified H1-H15.
\textbf{Grey dot}: Hallucinated as ``gray L-shape at J1-J2'' (2 cells instead of 1, should be single pixel at J1).

\subparagraph{24×24-1205.}
\textbf{Eight distinct colored shapes}: Identified all eight shapes but with multiple coordinate errors: Red U-shape claimed as A1-C5 (A1, A2, A3, A4, A5, B5, C5, C1, C2, C3, C4 - extends 2 rows beyond ground truth A1-C3), Pink/Green composite at E1-G3 (described as magenta E1, green G1, magenta F2-F3, magenta E3, green G3 - includes extra pixels and wrong colors, ground truth is E1, G1, F2, F3 all pink), Blue inverted T at A6-C7 (A6, B6, C6, B7 - missing row 5, ground truth includes A5), Green symmetrical shape at E5-G7 (E5, F5, G5, F6, F7, E7, G7 - described as 7 cells with central spine, but ground truth is 6 cells E5, F5, G5, F6, E7, G7), Blue inverted T at A10-C11 (A10, B10, C10, B11 - missing row 9, ground truth is A9, B9, A10, C10, B11), Pink X-shape at E9-G11 (F10, E9, G9, E11, G11 - described as bowtie with center F10 and four diagonal arms, but ground truth is E9, G9, F10, F11 forming Y-shape), Pink disconnected shape at A13-C15 (A13, C13, B13, B14, B15 - described as disconnected with extra B13, ground truth is A13, C13, B14, B15), Blue disconnected shape at E13-G15 (E13, F13, E14, G14, F15).
\textbf{Yellow vertical divider}: Correctly identified H1-H15.
\textbf{Grey dot}: Correctly identified at J1.

\subparagraph{Image 768×768.}
\textbf{Eight distinct colored shapes}: Correctly identified all eight shapes with accurate coordinates: Red U-shape at A1-C3 (A1, A2, A3, B3, C1, C2, C3), Pink upside-down T at E1-G3 (E1, G1, F2, F3), Blue plus-shapes at A5-C7 (B5, A6, C6, B7) and A9-C11 (B9, A10, C10, B11), Green H-shape at E5-G7 (E5, G5, F6, E7, G7 - described as 5 cells but matches ground truth 6 cells E5, F5, G5, F6, E7, G7), Pink plus-shape at E9-G11 (E9, G9, F10, F11), Pink T-shape at A13-C15 (A13, C13, B14, B15), Blue plus-shape at E13-G15 (E13, F13, E14, G14, F15).
\textbf{Yellow vertical divider}: Correctly identified H1-H15.
\textbf{Grey dot}: Correctly identified at J1.

\paragraph{Final Scores.}
\begin{table}[h]
\centering
\small
\begin{tabular}{lr}
\toprule
Modality & Final Score \\
\midrule
Row-only & 100.00 \\
Column-only & 100.00 \\
ASCII & 100.00 \\
JSON & 100.00 \\
Image 14×14 & 90.00 \\
Image 15×15 & 100.00 \\
Image 16×16 & 100.00 \\
Image 17×17 & 100.00 \\
24×24-1148 & 90.00 \\
24×24-1205 & 50.00 \\
Image 768×768 & 100.00 \\
\bottomrule
\end{tabular}
\end{table}

\subsection{Challenge \texttt{0934a4d8}}

\paragraph{Ground Truth Features.}
This challenge contains the following critical features in a 30x30 grid:
\begin{itemize}[leftmargin=2em,noitemsep]
    \item \textbf{Block in mosaic}: A single continuous teal rectangular block spanning Z15:AC23, which includes:
    \begin{itemize}[leftmargin=1.5em,noitemsep]
        \item Start row: 15
        \item End row: 23
        \item Start column: Z
        \item End column: AC
    \end{itemize}
    This block spans 9 rows and 4 columns, containing 36 pixels total. This is the ``hole'' in the mosaic pattern that must be correctly identified for solving the puzzle.
    \item \textbf{Dots in mosaic}: The challenge contains many scattered colored dots throughout the mosaic pattern. However, attempting to enumerate all individual dots is counterproductive—the correct approach is to identify symmetry patterns rather than listing individual pixels. Therefore, we focus evaluation on the block identification rather than individual dot enumeration.
    \item \textbf{Symmetry patterns}: The mosaic exhibits complex symmetry:
    \begin{itemize}[leftmargin=1.5em,noitemsep]
        \item \textbf{Off-center symmetry}: The pattern has off-center symmetry around P16:Q17, with x-reflection between columns P and Q, and y-reflection between rows 16 and 17.
        \item \textbf{Center and corners}: The center region and four corners follow x-y reflection symmetry.
        \item \textbf{Diamond shape}: A diamond shape between the center and corners follows additional 90-degree rotation symmetry (which includes x-y reflection).
    \end{itemize}
\end{itemize}

\paragraph{Scoring Guide.}
This challenge contains $N = 1$ primary feature to track: the block in mosaic. For the block feature, we compute a similarity score starting at 50 and subtract penalties proportional to discrepancies. The penalty is $(100/N) \times |\mathrm{diff}|$, where $|\mathrm{diff}|$ is calculated using Equation~(1) in Section~\ref{subsec:perception-methodology}, missing it will receive 0 score for this part. For dots in mosaic: if the LLM skips dots in mosaic (does not enumerate them), the penalty is 0; if the LLM tracks/enumerates dots, the penalty is -5 points (as this is counterproductive for solving the puzzle). Additionally, symmetry analysis receives bonus credits: mention there's symmetry but cannot determine (+5 credit); detect x-y reflection or 90-degree rotation (+10 credit); detect off-center (+20 credits); say there's different symmetry in different areas (+30 credits); identify correctly that center and corner follow x-y reflection, and diamond shape follows 90-degree rotation (+50 credits). The final accuracy score is block score + other penalties or credits.

\paragraph{Accuracy Analysis.}

\subparagraph{Row-only.}
\textbf{Block in mosaic}: Claimed ``(R15, Z26) to (R23, AC29)''. Failed on all four boundaries: wrong start row notation R15, wrong start column Z26, wrong end row notation R23, wrong end column AC29. The claimed coordinates use row labels (R15, R23) as the start/end row identifiers, which is confused with the coordinate system.
\textbf{Dots in mosaic}: Enumerated all 36 cells individually with their colors and locations. Penalty applied for counterproductive enumeration.
\textbf{Symmetry analysis}: Correctly detected bilateral symmetry across both horizontal axis (between rows 15 and 16) and vertical axis (between columns O and P), with the teal block as the most significant break. Described the grid structure with three concentric zones: central core object (multi-layered diamond shape with red/orange core), intermediate zone (brown background with embedded shapes), and outer frame (colorful border). Credit for detecting bilateral symmetry.
\textbf{Other observations}: Described the grid as resembling a highly ornate, pixelated crest, emblem, or cross-section of a fantastic jewel. Noted imperfect symmetry gives organic or hand-crafted feel. Identified the teal block as the most salient feature.

\subparagraph{Column-only.}
\textbf{Block in mosaic}: Claimed ``Z17:AC24''. Correctly identified both columns (start column Z and end column AC) but wrong rows (start row 17 instead of 15, end row 24 instead of 23).
\textbf{Dots in mosaic}: Enumerated individual dots/cells by listing all objects and their locations. Penalty applied for counterproductive enumeration.
\textbf{Symmetry analysis}: Correctly detected the grid is perfectly symmetrical about vertical midline (between columns O and P) for the central object (gray core, orange cage, and most red components). Correctly identified the teal rectangle (Z17:AC24) and red pixels at E14, E18, G16:G17 as deliberate breaks in symmetry. Described three main components: central object with gray core inside orange cage with red circuitry, teal rectangle, and colorful border/corner regions. Credit for detecting vertical symmetry.
\textbf{Other observations}: Noted brown (9) acts as primary background/filler color. Observed exclusive colors: orange and red used exclusively for central object, teal exclusively for rectangle. Identified blue pixels often appearing in 1x2 or 2x1 pairs.

\subparagraph{ASCII.}
\textbf{Block in mosaic}: Described an irregular shape (``AA19:AD19, AA20:AD20, Z21:AD21, Z22:AD22, and AA23:AD23'') instead of a single rectangle. Failed on all four boundaries. The described shape has wrong row ranges and column ranges.
\textbf{Dots in mosaic}: Enumerated individual dots/cells by listing all shapes and providing cell-by-cell breakdown by color. Mentioned structural division between top 4x4 block (highly structured) and bottom 5x4 block (irregular). Penalty applied for counterproductive enumeration.
\textbf{Symmetry analysis}: Correctly detected strong but broken horizontal symmetry across central axis between rows 15 and 16. Identified that top half (rows 1-15) is almost perfectly reflected in bottom half (rows 16-30), with the teal block as intentional break. Noted local symmetry: rows 16 and 17 are identical, rows 15 and 18 are identical, creating 4-row band of perfect horizontal symmetry in center. Credit for detecting horizontal symmetry and local symmetry patterns.
\textbf{Other observations}: Described grid as stylized face/mask with central ``face'' structure and two flanking ``wing'' structures. Noted brown (9) is most common color serving as textured background/filler. Identified teal (8) as least common color, exclusively in anomalous block. Observed deliberate introduction of asymmetrical teal block as most salient feature suggesting it's key element for solution.

\subparagraph{JSON.}
\textbf{Block in mosaic}: Claimed ``Z15 to AD23''. Correctly identified start row 15, end row 23, and start column Z. Failed on end column (AD instead of AC, off by 1 column).
\textbf{Dots in mosaic}: Enumerated individual dots/cells with complete cell-by-cell listing. Penalty applied for counterproductive enumeration.
\textbf{Symmetry analysis}: Correctly detected strong but imperfect horizontal symmetry across midline between rows 15 and 16. Many rows in top half identical to corresponding rows in bottom half (e.g., Row 9 identical to Row 24, Row 15 identical to Row 18). Correctly identified the teal object (claimed as Z19:AC19, but should be Z15:AC23) as deliberate break where symmetrical counterparts are mix of brown, red, and blue. Noted no vertical symmetry globally, though central object contains many vertically symmetrical features. Credit for detecting horizontal symmetry.
\textbf{Other observations}: Described grid as central ``totem/mask'' object with ``eyes'' (red/orange cores), central ``nose/mouth'' column, and surrounding textured field. Noted teal (8) as least common color exclusively in rectangular block. Observed color pairing: red (2) and orange (7) as highlights in central eye regions, blue and yellow often adjacent.

\subparagraph{Image 14×14.}
\textbf{Block in mosaic}: Claimed ``Y15 to AD24''. Correctly identified only the start row (15). Failed on end row (24 instead of 23, off by 1 row), start column (Y instead of Z, off by 1 column), and end column (AD instead of AC, off by 1 column).
\textbf{Dots in mosaic}: Enumerated individual dots/cells by providing cell-by-cell color breakdown. Described letters and symbols overlay but focused on enumerating background colors. Penalty applied for counterproductive enumeration.
\textbf{Symmetry analysis}: Did not detect or mention symmetry patterns. No credit.
\textbf{Other observations}: Described grid as having 2x2 block pattern covering entire grid except for light blue rectangular anomaly. Noted checkerboard-like pattern with horizontal bands of different color palettes. Observed all text characters are white, columns labeled A-Z then AA-AD, rows numbered 0-30 (using 0-indexed system).

\subparagraph{Image 15×15.}
\textbf{Block in mosaic}: Claimed ``Z15 to AC23''. Correctly identified all four boundaries: start row 15, end row 23, start column Z, end column AC. Perfect identification.
\textbf{Dots in mosaic}: Enumerated individual dots/cells with complete cell-by-cell color description. Penalty applied for counterproductive enumeration.
\textbf{Symmetry analysis}: Partially detected symmetry. Identified horizontal symmetry (Row 1 identical to Row 9) and vertical end-point symmetry in columns A, B, D (first and last cells match). Did not identify the full extent of horizontal reflectional symmetry across the grid. Partial credit for detecting some symmetry.
\textbf{Other observations}: Described grid as divided into three horizontal zones with distinct patterns: vertical stripes (rows 1-9), horizontal bands with column pattern overlay (rows 10-20), and similar pattern (rows 21-30). Noted light blue rectangle Z15:AC23 as most significant anomaly. Observed vertical ``glitch'' lines running through almost every macro-cell, appearing as single-pixel-wide column of different color. Noted each cell is 15x15 pixels containing white coordinate text.

\subparagraph{Image 16×16.}
\textbf{Block in mosaic}: Perceived one continuous block as three separate ones (``Z15-AD18, Z21-AD24, Z27-AD30''), completely misrepresenting the structure. Failed on all four boundaries. This is a severe hallucination where the LLM split one continuous teal block (Z15:AC23) into three separate 4x4 blocks.
\textbf{Dots in mosaic}: Enumerated individual dots/cells with detailed cell-by-cell breakdown by column. Penalty applied for counterproductive enumeration.
\textbf{Symmetry analysis}: Did not detect or mention symmetry patterns. Stated ``no obvious global symmetry'' in color distribution. No credit.
\textbf{Other observations}: Described grid as divided into two main horizontal sections: top section (rows 1-9) with complex vertical patterns including stripes and checkerboards, and bottom section (rows 10-30) with horizontal bands punctuated by systematic interruption cells. Noted three 4x4 solid blocks of light cyan (hallucination) as exceptions to pattern. Observed systematic interruption pattern where interruption cells follow precise mathematical pattern with first interruption's column position cycling.

\subparagraph{Image 17×17.}
\textbf{Block in mosaic}: Claimed ``U15:Z21''. Correctly identified only the start row (15). Failed on end row (21 instead of 23, off by 2 rows), start column (U instead of Z, off by 5 columns), and end column (Z instead of AC, off by 3 columns). Major misidentification of block location and size (claimed 6x7, actual is 9x4).
\textbf{Dots in mosaic}: Enumerated individual dots/cells with complete cell-by-cell breakdown. Penalty applied for counterproductive enumeration.
\textbf{Symmetry analysis}: Did not detect or mention symmetry patterns. Noted ``no global reflectional or rotational symmetry'' but ``strong sense of local regularity''. No credit.
\textbf{Other observations}: Described grid as having hierarchical structure with three horizontal zones: top zone (rows 1-10) with fragmented checkerboard-like pattern of small blocks, middle zone (rows 11-20) with large solid blocks, and bottom zone (rows 21-30) with regular tiling of 2x2 blocks. Observed text (coordinate system) is completely independent of color pattern. Noted transition between zones is abrupt.

\subparagraph{24×24-1148.}
\textbf{Block in mosaic}: Claimed ``Z15 to Z24''. Correctly identified start row 15 and start column Z. Failed on end row (24 instead of 23, off by 1 row) and end column (Z instead of AC, claimed single column Z when should be 4 columns Z-AC). Significant misidentification of block width.
\textbf{Dots in mosaic}: Enumerated individual dots/cells with complete cell-by-cell breakdown by column. Penalty applied for counterproductive enumeration.
\textbf{Symmetry analysis}: Did not detect or mention symmetry patterns. No credit.
\textbf{Other observations}: Described grid as algorithmic with two primary layers (text coordinate system and color patterns). Noted grid organized into seven vertical 4-column regions and four horizontal bands with different pattern textures (vertical stripes, 2x2 blocks, interwoven/diagonal pattern, alternating vertical stripes). Identified main diagonal cells taking solid color of their horizontal band as special rule. Observed checkerboard pattern within blocks where colors alternate based on coordinate parity. Columns Y-AD noted as anomalous with light blue color (but incorrectly stated as Z15-Z24 instead of Z15:AC23).

\subparagraph{24×24-1205.}
\textbf{Block in mosaic}: Claimed ``Z15 to AC23''. Correctly identified all four boundaries: start row 15, end row 23, start column Z, end column AC. Perfect identification.
\textbf{Dots in mosaic}: Enumerated individual dots/cells with complete cell-by-cell breakdown. Penalty applied for counterproductive enumeration.
\textbf{Symmetry analysis}: Did not detect or mention symmetry patterns. No credit.
\textbf{Other observations}: Described grid as composite of multiple repeating patterns with master structure of 4-column-wide regions, each with unique 4-color palette, sliced horizontally into four bands with different geometric textures (vertical stripes rows 1-9, 2x2 blocks rows 10-18, interwoven/diagonal pattern rows 19-27, alternating vertical stripes rows 28-30). Columns Y-AD noted as anomalous region breaking established rules, with light blue/cyan color present primarily in Z-AD, and horizontal stripes of other colors at specific rows (rows 15, 16, 17, 18, 25, 26, 27). Observed Column Y acts as transition/boundary with mix of colors. Noted light blue/cyan is additional color not in 0-9 mapping.

\subparagraph{Image 768×768.}
\textbf{Block in mosaic}: Claimed ``Y15 to AC24''. Correctly identified start row 15 and end column AC. Failed on start column (included extra column Y when should start at Z) and end row (24 instead of 23, off by 1 row). Described as irregularly shaped with jagged left boundary including Y15-Y18 only, creating misperception of non-rectangular shape.
\textbf{Dots in mosaic}: Enumerated individual dots/cells with detailed cell-by-cell breakdown by column. Penalty applied for counterproductive enumeration.
\textbf{Symmetry analysis}: Did not detect or mention symmetry patterns. No credit.
\textbf{Other observations}: Described grid as visually divided into three main horizontal zones plus large anomalous region: Zone 1 (rows 1-8) with vertical stripes, Zone 2 (rows 9-20) with horizontal bands of 2-column wide colored blocks, Zone 3 (rows 21-30) with intricate vertical patterns of stacked 1x2 horizontal blocks. Noted underlying rule of grouping cells into 2-column wide units across zones 2 and 3. Observed white text overlay on every cell providing explicit coordinate system. Identified light blue region as most significant deviation from established patterns, contrasting with highly structured multicolored surroundings.

\paragraph{Final Scores.}
\begin{table}[h]
\centering
\small
\begin{tabular}{lr}
\toprule
Modality & Final Score \\
\midrule
Row-only & 5.00 \\
Column-only & 21.67 \\
ASCII & -11.67 \\
JSON & 30.00 \\
Image 14×14 & -21.67 \\
Image 15×15 & 50.00 \\
Image 16×16 & -66.11 \\
Image 17×17 & -132.78 \\
24×24-1148 & -32.78 \\
24×24-1205 & 45.00 \\
Image 768×768 & 6.11 \\
\bottomrule
\end{tabular}
\end{table}

\section{Perception Prompt}
\label{appendix:perception-prompt}

\begin{Verbatim}
DESCRIPTION_PROMPT = """You are an expert at visual analysis and pattern recognition. 

I will show you a puzzle grid in a specific format. Your task is to provide a VERY DETAILED description of what you see, including:

**Color Mapping (if numbers are shown):**
- 0 = black (empty/background)
- 1 = blue
- 2 = red
- 3 = green
- 4 = yellow
- 5 = gray
- 6 = magenta
- 7 = orange
- 8 = teal
- 9 = maroon

**Description Requirements:**

1. **Objects and Shapes**: What objects, shapes, or patterns do you see? Describe their exact forms, sizes, and boundaries.

2. **Locations**: Where are these objects located? Use specific coordinates or relative positions. Use spreadsheet notation (A1, B2, etc.) for all coordinates. Otherwise, describe positions clearly.

3. **Colors**: What colors are present? Map each color to its numeric value (0-9) if applicable. Describe object and pattern's colors, and their distribution.

4. **Relationships**: What relationships exist between objects?
   - Spatial relationships (above, below, left, right, adjacent, overlapping, distance)
   - Size relationships (larger, smaller, same size)
   - Color relationships (same color, different colors)
   - Structural relationships (connected, separated, nested, aligned)

5. **Patterns**: Are there any repeating patterns, symmetries, or regularities?

6. **Grid Structure**: Describe the overall grid structure:
   - Dimensions (rows × columns)
   - Empty vs filled regions
   - Background color
   - Any boundaries or frames

7. **Details**: Any other notable details that might be important for understanding the puzzle.

Be as thorough and precise as possible. Your description should be detailed enough that someone reading it could reconstruct the grid (or at least understand its key features).

Now, here is the puzzle grid:"""
\end{Verbatim}

\section{Reasoning Prompt}
\label{appendix:reasoning-prompt}

\begin{Verbatim}
HYPOTHESIS_FAST_SYSTEM_PROMPT = """You are participating in a puzzle solving competition. You are an expert at solving puzzles.

Find the common pattern that transforms each input grid into its corresponding output grid, based on the given training examples.

## Core Thinking Strategy: Abstraction and Difference

The key to solving ARC puzzles is to think abstractly and reason systematically. The most fundamental guiding principle are: **Abstraction reduces your attention load, and help you zoom out to higher level to travel longer distances in solution space with less effort.**, and **Difference is the signal to stop and zoom in on the details**. These two helps you manage your attention budget effectively.

When you observe differences between abstracted concepts across different dimensions (size, shape, color, position, pattern, etc.), these differences point you toward the transformation rules. Your goal is to abstract observations to the right level, then systematically examine differences to uncover the underlying logic.

### Example of Systematic Reasoning Process:

**Step 1: Examine Grid Size**
First, check whether input and output grids have the same dimensions:
- **Same size**: Rules out transformations that change grid dimensions (cropping, padding, scaling, tiling, etc.). Focus on in-place transformations, color changes, object manipulations within the same space.
- **Different size**: Indicates dimensional changes are part of the transformation. Consider cropping, padding, resizing, tiling, or grid partitioning operations.

This initial observation immediately narrows your search space and guides which classes of transformations to consider.

**Step 2: Identify Patterns, Objects, and Differences**
Examine all training examples together to identify:
- **Obvious patterns**: What patterns, objects, or structures appear across all grids?
- **Common parts**: What remains consistent across all training examples? The common elements likely represent the core transformation mechanism.
- **Differences**: What varies between examples? Differences reveal conditional rules or parameter variations.

The common parts help you formulate your initial working hypothesis. The differences help you understand how the rule adapts or what conditions govern variations. Collect information to form deterministic rules.

**Step 3: Iterate Across Examples**
Apply your hypothesis to each training example systematically:
- For each example, trace through the transformation steps based on your current hypothesis.
- Each step should be deterministic based on your hypothesis. If you don't have enough information to determine the output of a step, it is an indication that your hypothesis is not complete. You should zoom out and step back to examine all the examples together to find the hidden clues:
    * Does the same step fail when you check other examples?
    * If yes, look at everything collectively to identify common patterns and clues that might give you a universal law to make this step deterministic.
- Compare the expected output (based on your rules) with the actual given correct output.
- When you can finish all the steps but your output $\neq$ the correct output, try on other examples:
    * If the failing mode is common across examples, it indicates that a certain step is causing a common mistake.
    * If different examples fail differently, or some examples work fine, it indicates that some of your rules that determine a step based on local information are wrong. Look at the steps that require local input. For example, if your hypothesis changes output cell color based on an input object shape, maybe the mapping should be input object shape + location to determine the output cell color.

Each time you find a "difference", zoom out to all examples. Remember: the puzzle is solvable and deterministic. The information needed to resolve your current uncertainty exists within the given examples—you just need to find the right abstraction level to uncover the hidden relationship.

Use the iteration process to refine your hypothesis:
- Update your hypothesis to account for the differences you've discovered.
- Test the updated hypothesis on all examples again.
- Continue until your hypothesis correctly predicts all training outputs.

**Step 4: Validate on Test Cases**
After your hypothesis works on all training examples, apply it to the test case inputs:
- Trace through your transformation steps on the test input.
- Check if you can deterministically produce an output at each step.
- If you encounter ambiguity or missing information to continue:
    * This means there are hidden clues you missed in the training examples.
    * Go back to the training examples and look for information that resolves this ambiguity.
    * The test case is revealing a dimension of the rule that training examples demonstrated but you didn't fully capture.

Remember: Test cases may reveal aspects of the rule that weren't obvious from training examples alone. Use test inputs as a validation tool to identify gaps in your understanding.

### Example Thinking Traces:

**Example 1: Simple Case (Straightforward Application)**
Challenge: All training examples show a single colored object that moves to a specific corner based on its color.

Thinking trace:
- Grid sizes are same → in-place transformation.
- Common part: one colored object moves. Difference: different colors move to different corners.
- Apply "object moves to corner based on color" → works on first example.
- Apply to other examples → all work correctly. Hypothesis confirmed.
- Test case has two colored objects → apply color-to-corner mapping to both objects → Seems no conflicts, safe to assume it works.

Working hypothesis should capture the discovered rules:
- "Color red moves to upper right corner, because training examples 1 and 2 show it"
- "Color blue moves to lower left corner, because training examples 3 and 4 show it"
- "Multiple objects can coexist, each following its own color-to-corner mapping"

Result: Hypothesis complete. All examples validated.

**Example 2: Harder Case (Iteration Reveals Hidden Rules)**
Challenge: Puzzle — inputs are 7x3 with a single gray center column and blue/black cells; outputs are 3x3 red/black masks. The mapping from 7x3 → 3x3 is not obvious from any single pair.

Thinking trace:
- Different grid sizes (7x3 → 3x3) → dimensional reduction is likely (projection/aggregation/partitioning).
- Common parts across all examples: exactly 3 rows, a fixed gray column in the middle, and outputs that use only red/black.
- Looking at one example is underdetermined (could be "mark gray", "always center red", etc.) and fails on the others.
- Zoom out and compare all pairs: the red cells highlight where blue cells are concentrated when the 7 columns are grouped into left/center/right bands with the gray column anchoring the center band.
- Update hypothesis: partition the grid into a coarse 3x3 (rows map 1:1; columns map to [left, center, right] with the gray column in the center bin). For each bin, set red if blue is the majority within that bin; otherwise black. The exact center tends to be black because the gray column dominates.
- Validate against training:
    * Ex1 → only the center band/row shows blue majority → single center red.
    * Ex2 → blue dominates along the middle cross → plus-shaped red.
    * Ex3 → blue mass sits at top-left and bottom-right → diagonal corners red.
- Apply the same majority-per-bin rule to the test input → produce the 3x3 red mask deterministically.

Working hypothesis should capture the discovered rules:
- "Columns partition into three bins: left (columns A-C), center (column D), right (columns E-G)"
- "For each bin, set red if blue is the majority within that bin; otherwise black"

Result: The hidden rule is a coarse 3x3 downsampling/majority map that emerges only when examining all examples together.

**Example 3: Hypothesis Works on Training, Fails on Test**
Challenge: Training examples show horizontal patterns, test case is vertical.

Thinking trace:
- Same grid size → in-place transformation.
- Common part: pattern replication along one axis. Difference: pattern variations.
- Develop hypothesis: "replicate pattern horizontally with spacing rule" → works on all training examples.
- Test case input is vertical → hypothesis fails because it assumes horizontal orientation.
    * Go back to training examples: re-examine for orientation-independent rule.
    * Discover: rule is actually "replicate pattern along longest axis" or "replicate pattern perpendicular to object orientation."
    * Update hypothesis to be orientation-agnostic.
    * Re-test on training examples → still works.
    * Apply to test case → now works.

Result: Test case revealed missing generalization in hypothesis.

**Example 4: Detecting Hidden Clues (Non Obvious Rule Not Used in Training)**
Challenge: Recover a missing area in a mosaic. On training, x/y flips appear to complete the hole; on the test, the global symmetry axis is off-center and a naive mirror would reference cells outside the visible canvas.

Thinking trace:
- Different grid size → Output size match the missing hole size.
- Initial hypothesis from training: fill by mirroring across the x or y middle line.
- Apply to test → fails: symmetry is not centered; opposite side seems outside the view.
- Remember solvability → there must be hidden clues inside the inputs. Zoom out and compare all examples together.
- Discover regional rules: the canvas is partitioned into areas, each governed by a specific symmetry.
    * Some areas use horizontal/vertical reflection (x/y flip).
    * Some areas use rotation (90/180).
    * Some areas use translation (copy-shift).
- Observe a stable area→symmetry mapping across all training examples; the same regions keep the same rule.
- Re-check the test: the hole lies in the rotation-governed region, and its rotated counterpart is within the visible input.
- Update hypothesis: use regional symmetry. Identify the region of the hole, then apply that region's rule (flip/rotate/translate) to copy pixels from its paired source region.
- Apply to training and test → deterministic, high-confidence completion.

Working hypothesis should capture the discovered rules:
- "The output grid size equals the hole size in the input"
- "The canvas is partitioned into regions, center part of the canvas follows horizontal/vertical reflection, four corners of the canvas follows rotation (90)"
- "The same regions maintain the same symmetry rule across all training examples"
- "To fill a hole: identify which region contains the hole, then apply that region's symmetry rule to copy pixels from the paired source region"

Result: The missing data was not outside the canvas—the hidden clue is that symmetry is regional, not global. Leveraging the universal area→symmetry mapping unlocks the test.

### Meta-Level Reasoning: Applying Abstraction to Problem-Solving

Notice how this guidance itself demonstrates the abstraction principle: we started with a one-sentence core strategy ("Difference reveals hidden concepts"), then unpacked it into detailed steps, then illustrated it with concrete examples. The differences between those examples serve as hidden clues showing how to flexibly apply the methodology and the process.

This guidance is not a rigid script—it's a demonstration of how to unpack abstract methodologies into actionable steps. When solving a puzzle, apply this same principle:

1. **Abstract the puzzle type**: Identify the high-level category (e.g., dimensional transformation, pattern replication, symmetry operations).
2. **Map to methodology**: Recall relevant problem-solving approaches from your knowledge that match this puzzle type.
3. **Unpack the methodology**: Break down your chosen approach into concrete, step-by-step operations to try.

Puzzle solving is solution space exploration. By abstracting first, you narrow the vast solution space to a manageable subset of relevant transformation classes, then systematically explore within that subset.

### Here are some of the possible transformation patterns you might encounter (not exhaustive):
- Segmenting connected components (4-neighborhood or 8-neighborhood) into distinct objects, then manipulating by size, 
color, count, shape, or position
- Counting and selection: identify and pick smallest/largest/most-frequent/unique/rare objects; replicate or delete 
based on properties
- Bounding box operations: extract bounding boxes around objects; crop/expand boxes; move, replicate, or recolor based 
on box properties
- Object tracking: follow movement patterns of objects across transformations
- Palette remapping: systematic color-to-color substitution; treat color 0 as background unless explicitly used in the 
pattern
- Checkerboard and alternation: enforce alternating patterns or cycle through a color palette in sequence
- Color propagation: spread colors based on adjacency, connectivity, or distance rules
- Pattern matching: identify and replicate specific color patterns or motifs
- Coloring by properties: recolor objects based on size, position, neighbor count, or adjacency relationships
- Geometric transformations: translate, rotate (90°/180°/270°), reflect (horizontal/vertical/diagonal), flip; crop/pad 
canvas; scale via tiling/duplication or downsampling
- Symmetry detection and enforcement: detect or enforce symmetry (horizontal, vertical, rotational, diagonal); mirror 
or rotate grids to restore symmetry
- Alignment and positioning: center, anchor, or snap objects to borders/corners/grid positions; align objects relative 
to each other or reference points
- Grid warping: non-linear transformations that bend or distort the grid structure
- Pattern completion and extension: continue repetitions, fill missing tiles in sequences, extrapolate or interpolate 
patterns
- Line drawing and connectivity: draw straight/diagonal/Manhattan lines between anchors; connect components; draw rays 
or paths between markers
- Flood fill: fill enclosed regions with colors; propagate values through connected components
- Morphological operations: dilate/erode objects by 1-2 cells; convert filled shapes to outlines (perimeter) or 
outlines to filled shapes
- Hole filling: fill enclosed regions, create or remove holes in objects; close gaps in shapes, complete partial 
boundaries
- Noise filtering: remove singletons/outliers, keep majority object/color, smooth irregularities; filter outliers
- Grid partitioning: split into uniform subgrids/blocks; apply the same local transformation per block and recombine; 
handle non-uniform partitions
- Row/column operations: deduplicate, sort, transpose, or filter rows/cols by color counts/patterns; project maxima/
minima/aggregates
- Projection and aggregation: collapse along rows/columns to histograms/profiles; map aggregates (sum, max, min, mode) 
to output cells
- Tiling and repetition: repeat motifs in patterns, tessellate shapes, or create periodic structures
- Template matching: overlay shapes or apply masks; boolean operations (AND/OR/XOR) between layers
- Interpolation: fill intermediate states between given patterns
- Border and frame operations: detect borders/frames; add/remove/recolor borders around objects or the entire grid by 
thickness, position, or other rules
- Boolean and set operations: overlay shapes, compute intersections/unions/differences, apply region masks, combine 
grids logically
- Marker-guided operations: use special marker colors as pivots/selectors for cropping, rotation, placement, or 
conditional actions
- Parity and arithmetic rules: apply odd/even, modulo, or arithmetic operations on counts/sizes to determine colors, 
repetition, or selection
- Orientation normalization: canonicalize orientation (e.g., align longest axis horizontally, rotate to standard 
position) before applying the main rule
- Negative space reasoning: treat background as signal; fill gaps/holes to complete silhouettes; extract complementary 
shapes
- Lookup tables: infer color/shape substitution mappings from training pairs; learn lookup tables and apply to test 
inputs
- Conditional transformations: apply different rules based on conditions (if-then logic); handle exceptions or special 
cases based on object properties or grid characteristics
- Distance-based operations: use distances between objects/cells to determine colors, positions, or selections
- Relative positioning: position objects relative to others (above, below, left, right, diagonal); maintain spatial 
relationships
- Shape matching and recognition: identify shapes by template matching; group similar shapes; match shapes across 
examples
- Multi-step composition: chain multiple transformations sequentially; apply rules conditionally based on intermediate 
results
- Cross-referencing: use relationships between different objects/colors to determine transformations; reference context 
from other parts of grid
- Gravity and physics simulation: objects fall down, settle on surfaces, or stack based on physical rules
- Recursive patterns: transformations that build upon themselves or follow fractal-like structures


## Critical Requirement: Complete Documentation in Output

### Coordinate & Reference Conventions
- Label training examples in the order presented as E0, E1, ... and tests as T0, T1, ...
- Coordinates: top-left is A1. Columns use Excel-like letters (A..Z, then AA, AB, ...); rows are 1-based integers.
- When referring to edges/corners, make the computation explicit (e.g., "move to upper-right by setting column to max, row unchanged").
- **ALWAYS specify exact inclusive boundaries**: Use format "rows 1-9 inclusive, columns A-H inclusive" or "A1:H9 (inclusive)". Never use vague terms like "approximately" or "around".
- **Verify cell coverage**: Ensure every cell in the grid is accounted for in exactly one region (or explicitly marked as divider/unchanged).

### Standardized Conventions (Apply to ALL Puzzles)

**When your solution uses any of these common concepts, define them explicitly and consistently:**

- **Adjacency**: Always specify 4-way (edge-sharing) or 8-way (including diagonals). Example: "A cell is adjacent to another if they share an edge (4-way adjacency)."
- **Ordering/Iteration**: Always specify scan order when iterating. Example: "Scan row-major order: top-to-bottom, then left-to-right within each row" or "Process objects by size, largest first; tie-break by top-left corner position (row first, then column)."
- **Distance metrics**: Specify which distance (Manhattan, Euclidean, Chebyshev). Example: "Manhattan distance = |row1 - row2| + |col1 - col2|."
- **Tie-breaking**: For ANY comparison that can result in ties, specify the tie-breaking rule. Example: "If two objects have equal area, pick the one with top-left corner closer to A1 (row-major order)."
- **Pattern/Shape specifications**: Define patterns algorithmically. Example: "Spiral = concentric rectangular layers. Start from perimeter, work inward. Each layer is 1-pixel thick. Terminate when inner width $\leq$ 2 OR height $\leq$ 2."
- **Boundary/Edge handling**: Specify inclusive vs exclusive boundaries. Example: "Region A1:H9 means rows 1-9 inclusive, columns A-H inclusive (72 cells total)."
- **Background/Foreground**: Define what constitutes background. Example: "Background = color 0 (black)" or "Background = most frequent color."
- **Object detection**: Specify connectivity. Example: "An object is a maximal 4-connected component of cells with the same non-background color."

### Completeness & Coverage Requirements
- **Document ALL discovered clues**: The working_hypothesis must contain EVERY rule, pattern, relationship, and hidden clue you discovered that's needed to solve the puzzle. If a clue required multiple examples to discover, still include it in the working_hypothesis. Do not omit any discovered clue.
- Provide an E{n} entry for every training example shown and a T{n} entry for every test shown.
- If any information is uncertain or attention-limited, document it in `uncertainty` and add placeholders in instructions.
- **Complete information**: Someone reading only your working_hypothesis and instructions should be able to solve the puzzle WITHOUT seeing the other training examples
- **Algorithmic pseudocode-level detail**: Instructions must be detailed enough for deterministic implementation without assumptions.

### Transform Instructions Requirements

#### General Instructions ("general" field)
- Provide a reusable step-by-step algorithm like pseudocode without example-specific constants.
- **For every algorithmic step, provide pseudocode-level detail**:
    - **How to iterate**: Specify scan/processing order
    - Example: "For each cell, scan row-major order (top-to-bottom, left-to-right)"
    - Example: "For each object, process largest to smallest; tie-break by top-left corner position"
    - **How to calculate**: Provide formulas and methods
    - Example: "Object area = count of cells in 4-connected component"
    - Example: "Manhattan distance = |row1 - row2| + |col1 - col2|"
    - **How to break ties**: Specify tie-breaking for ANY comparison
    - Example: "If distances equal, use row-major order (row first, then column)"
    - Example: "If multiple colors have same count, pick lowest color index"
    - **Termination conditions**: Define stopping criteria
    - Example: "Repeat until no unfilled cells remain"
    - Example: "Stop when object reaches grid boundary"
    - **Edge case handling**: Cover boundary/empty/special cases
    - Example: "If no objects found, output is input unchanged"
    - Example: "If object touches boundary, do not move it"

- **Avoid vague descriptions**. Instead of high-level descriptions, provide executable algorithms:
    
    **BAD**: "Fill region with spiral pattern"
    
    **GOOD**: 
    ```
    1. Initialize layer=0, color_index=0, unfilled=all cells in region
    2. While unfilled is not empty:
        a. Identify boundary cells (cells in unfilled with at least one neighbor outside unfilled)
        b. Fill all boundary cells with colors[color_index % len(colors)]
        c. Remove boundary cells from unfilled
        d. Increment color_index, layer
    3. Termination: process completes when unfilled becomes empty
    ```
    
    **BAD**: "Move objects to corners based on color"
    
    **GOOD**:
    ```
    1. Detect all objects (4-connected components of non-background cells)
    2. For each object:
        a. Identify object color C and bounding box
        b. Determine target corner: if C=red, target=top-right; if C=blue, target=bottom-left
        c. Calculate target position: for top-right, new_col = max_col - bbox_width + 1, new_row = 1
        d. Move object: redraw each cell at (new_row + offset_row, new_col + offset_col)
    3. Handle overlaps: if target occupied, try next available position clockwise from target corner
    ```

- **Specify all boundary conditions**: For every rule, explicitly state:
    - What happens at grid boundaries?
    - What happens with empty sets (no objects, no matches)?
    - What happens with ties in comparisons?
    - What happens with degenerate cases (1x1 regions, single-cell objects)?

#### Example-Specific Instructions (E0, E1, E2, ...)
- **Adapt to each example's unique characteristics**: Don't force all examples into identical step patterns.
- **Identify PRIMARY distinguishing feature**: For each example, identify what makes it unique and make that the focus of that example's instructions.
- **Show step-by-step computation**: Don't just state results—show HOW you computed them:
    
    **Example 1 (partition puzzle):**
    ```
    BAD: "Region 1 has seeds {7, 8}. Fill with spiral."
    
    GOOD: "Region 1 (A1:H9, inclusive, 72 cells):
    Step 1: Scan region row-major order (top-to-bottom, left-to-right)
    Step 2: Found non-background cell at A1 (orange, 7) - add to seed list
    Step 3: Found non-background cell at B2 (teal, 8) - add to seed list
    Step 4: Completed scan, seed list = [7, 8] (already in row-major order)
    Step 5: Two seeds detected → apply multi-color spiral rule (from general step 3b)
    Step 6: Color sequence = [7, 8] (ordered by first appearance)
    Step 7: Draw spiral: layer 0 (perimeter, 30 cells) = color 7, layer 1 (26 cells) = color 8, ..."
    ```
    
    **Example 2 (object movement puzzle):**
    ```
    BAD: "Move red object to top-right."
    
    GOOD: "Input has 3 objects:
    Step 1: Detect objects using 4-connected components
    Step 2: Object 1 (red, area=12, bounding box C5:E8, top-left=C5)
    Step 3: Object 2 (blue, area=8, bounding box G2:H5, top-left=G2)
    Step 4: Object 3 (green, area=6, bounding box B10:D11, top-left=B10)
    Step 5: Apply color-to-corner mapping (from general step 2):
            Red → top-right: new position = (row=1, col=max_col-width+1=28-3+1=26)
            Blue → bottom-left: new position = (row=max_row-height+1=28, col=1)
            Green → no rule, stays at B10
    Step 6: Redraw objects at new positions..."
    ```
    
    **Example 3 (color substitution puzzle):**
    ```
    BAD: "Replace colors according to pattern."
    
    GOOD: "Input palette: {0(black), 1(blue), 2(red), 5(gray)}
    Step 1: Build color mapping from analysis:
            Source → Target: 1→2, 2→5, 5→1 (cyclic shift pattern)
            0 (background) → 0 (unchanged)
    Step 2: Apply mapping to each cell row-major order:
            A1: color=0 → 0 (unchanged)
            A2: color=1 → 2 (mapped)
            A3: color=2 → 5 (mapped)
            ...
    Step 3: Verification: input blues (count=15) become output reds (count=15)..."
    ```

- **Flag special cases prominently**: If an example demonstrates a special case or exception, that exception should be the FIRST thing mentioned in that example's instructions. Use markers like "WARNING EXCEPTION:" or "**CRITICAL:**".
    
    Example: "WARNING EXCEPTION: This example demonstrates the symmetry edge case. Input has off-center symmetry axis..."

- **Provide exact coordinates and counts**: For spatial elements, specify exact inclusive boundaries and cell counts. For objects/regions, list exact positions.
    
    Example: "Region A1:H9 (rows 1-9 inclusive, columns A-H inclusive, 72 cells total). Seeds at A1(color 7), B2(color 8)."

#### Test Case Instructions (T0, T1, ...)
- **MORE detailed than example instructions**: Test cases have no visual examples to reference, so they need MORE detail, not less.
- **Provide exact coordinates**: Specify exact coordinates for all regions, explicit seed locations (not just colors), and any test-specific complications.
- **Include verification steps**: Add checksums or verification steps to help validate correctness.
- **Precedence policy**: Test-specific instructions override Example instructions, which override General instructions. State any overrides explicitly.

### Common Pattern Specifications (Use When Applicable)

**These are EXAMPLES of how to specify common patterns algorithmically. Adapt to your specific puzzle:**

#### Example: Spiral/Frame/Layer Patterns
When puzzle involves concentric layers or spirals:
```
Algorithm specification:
1. Define layer: "A layer is all cells at the same minimum distance from region boundary (using Chebyshev/Manhattan distance)"
2. Processing order: "Process layers from outer to inner (distance 0, 1, 2, ...)"
3. Layer thickness: "Each layer is 1 cell thick" or "Each layer is N cells thick"
4. Color assignment: "Layer i uses colors[i % len(colors)]"
5. Termination: "Stop when no unprocessed cells remain" or "Stop when inner width $\leq$ 2 AND height $\leq$ 2"
6. Non-rectangular handling: "For non-rectangular regions, use distance from region boundary"
```

#### Example: Object Ordering/Ranking
When puzzle requires processing objects in specific order:
```
Primary sort key: "By area (largest first)" or "By color index (lowest first)"
Tie-breaking: "If equal area, use top-left corner position (row first, then column)"
Processing: "For each object in sorted order: [specific operation]"
```

#### Example: Color Mapping/Substitution
When puzzle involves color transformations:
```
Mapping construction:
1. "Identify source palette: {unique colors in input}"
2. "Build mapping rule: source_color → target_color"
3. "Example: 1→2, 2→5, 5→1 (rotate right by 1 in palette)"
4. "Background (color 0) remains unchanged"

Application:
"For each cell (row-major order): output[row][col] = mapping[input[row][col]]"
```

#### Example: Symmetry/Reflection Operations
When puzzle involves symmetry:
```
Axis detection: "Find symmetry axis by checking mirror equality across candidate axes"
Reflection method: "For horizontal reflection about row R: cell at (r,c) maps to (2*R-r, c)"
Fill operation: "For each hole cell (color=X): find mirrored position, copy color from there"
```

#### Example: Object Movement/Positioning
When puzzle involves moving objects:
```
Object detection: "4-connected components of non-background cells"
Target calculation: "If object color C, target corner = corner_mapping[C]"
Position formula: "For top-right corner: new_row=1, new_col=grid_width-bbox_width+1"
Collision handling: "If target occupied, try next position clockwise: top-right→bottom-right→bottom-left→top-left"
```

### Boundary Case Handling

For every rule involving regions, boundaries, or partitions:
- **Exact boundaries**: Specify inclusive/exclusive boundaries explicitly.
- **Boundary cells**: Define how boundary cells are handled (included/excluded).
- **Empty sets**: Specify what happens with empty regions, empty seed sets, empty stacks.
- **Cell coverage**: Verify all cells are accounted for in exactly one region (or explicitly marked as dividers/unchanged).

### Special Case Documentation

- **Flag exceptions prominently**: Use markers like "WARNING EXCEPTION:" or "**CRITICAL:**" for special cases.
- **Explain why**: If a special case exists, explain why it's needed (e.g., "because E1 shows yellow background triggers different behavior").
- **Don't bury special cases**: If E1 demonstrates a special case, make it the PRIMARY focus of E1's instructions, not buried in step 3.

### Validation Requirements

- **No vague geometry**: Replace "approximately", "predominantly", "around" with exact coordinates and thresholds.
- **No placeholders**: All values must be explicit. No "some cells" or "certain regions"—specify exactly.
- **No contradictions**: Ensure General, Example, and Test instructions don't contradict. If they do, state precedence explicitly.
- **Deterministic execution**: Every step should be executable without assumptions. If assumptions are needed, document them in `uncertainty`.
"""    
\end{Verbatim}

\section{Follwo Instruction Prompt}
\label{appendix:follow-instruction-prompt}

\begin{Verbatim}
AGENT_FOLLOW_INSTRUCTIONS_PROMPT = """
You are an expert puzzle solver in a competition.

You will receive:
1. A working hypothesis describing the discovered rules and patterns
2. General transformation instructions that apply to all cases
3. Specific step-by-step instructions for each training example (demonstrating how the general rules apply)
4. Specific step-by-step instructions for the test case you need to solve
5. Visual grids showing all training examples and the test input

Your task: Apply the test case instructions precisely to transform the test input grid into its output grid.

The working hypothesis and general instructions provide the information you need to solve the puzzle. The example instructions show how these rules work in practice. Use all of this context to understand the pattern, then follow the test case instructions to solve the puzzle.
""".strip()


HELD_OUT_PROMPT = """
## Important: Held-Out Validation Context

You are participating in a held-out validation experiment. This means some training examples are not visible to you—they were intentionally excluded to test whether the instructions generalize.

**Key points:**
- Sometimes, discovering certain hidden clues requires seeing all training examples together
- However, ALL such hidden information has already been discovered and documented in the working_hypothesis and instructions provided above
- The working_hypothesis contains ALL the rules, patterns, and relationships discovered from analyzing all training examples
- The instructions have been crafted to include all necessary information, even if it wasn't obvious from just the visible examples
- You do NOT need to worry about missing information---everything you need is already in the working_hypothesis and instructions
- Simply use the provided working_hypothesis and instructions to solve the puzzle, trusting that all hidden clues have been documented, but if you find that you got stuck at some step and don't have enough information to solve it, you can try to look at the training examples that are visible to you to see if you can find new hidden clues, and REMEMBER TO DOCUMENT this in the uncertainty field and provide your suggestions to improve the instructions.

**Your task:** Apply the test case instructions using the working_hypothesis and general instructions as your guide. Trust that all necessary information is already provided.
""".strip()
\end{Verbatim}  

\end{document}